%% file: HIN.tex
\documentclass[journal,compsoc]{IEEEtran}

\usepackage{paralist}
\usepackage{algorithm,algorithmicx}
\usepackage{amsthm}

\theoremstyle{definition}
\newtheorem{definition}{Definition}
\usepackage{mathrsfs}
\usepackage{algpseudocode}
\usepackage{multirow}
\usepackage{makecell}
\usepackage{xpatch}
\usepackage{enumitem}

\xpatchcmd{\algorithmic}{\itemsep\z@}{\itemsep=0.5ex}{}{}
\ifCLASSOPTIONcompsoc
  \usepackage[nocompress]{cite}
\else
  \usepackage{cite}
\fi
\ifCLASSINFOpdf
   \usepackage[pdftex]{graphicx}
   \graphicspath{{Figure/pdf/}}
\else
   \usepackage[dvips]{graphicx}
\fi
\usepackage{amsmath}
\usepackage{array}
\usepackage{url}

\begin{document}
%

\title{mSHINE: A Multiple-meta-paths Simultaneous Learning Framework for Heterogeneous Information Network Embedding}
%
%
%
%

\author{Xinyi Zhang, 
       Lihui Chen,~\IEEEmembership{Senior Member,~IEEE}
\IEEEcompsocitemizethanks{\IEEEcompsocthanksitem Xinyi Zhang and Lihui Chen are with the School of Electrical and Electronic Engineering, Nanyang Technological University, Singapore 639798, Singapore, E-mail: \{xinyi001@e.ntu.edu.sg, elhchen@ntu.edu.sg\}.}%
}

%
%

\markboth{IEEE Transactions on Knowledge and Data Engineering, ~Vol.~X, No.~X, FEBRUARY 2020}{XINYI \MakeLowercase{\textit{et al.}}: mSHINE: A Multiple-meta-paths Simultaneous Learning Framework for Heterogeneous Information Network Embedding}
\IEEEtitleabstractindextext{%
\begin{abstract}
 Heterogeneous information networks(HINs) become popular in recent years for its strong capability of modelling objects with abundant information using explicit network structure. Network embedding has been proved as an effective method to convert information networks into lower-dimensional space, whereas the core information can be well preserved. However, traditional network embedding algorithms are sub-optimal in capturing rich while potentially incompatible semantics provided by HINs. To address this issue, a novel meta-path-based HIN representation learning framework named mSHINE is designed to simultaneously learn multiple node representations for different meta-paths. More specifically, one representation learning module inspired by the RNN structure is developed and multiple node representations can be learned simultaneously, where each representation is associated with one respective meta-path. By measuring the relevance between nodes with the designed objective function, the learned module can be applied in downstream link prediction tasks. A set of criteria for selecting initial meta-paths is proposed as the other module in mSHINE which is important to reduce the optimal meta-path selection cost when no prior knowledge of suitable meta-paths is available. To corroborate the effectiveness of mSHINE, extensive experimental studies including node classification and link prediction are conducted on five real-world datasets. The results demonstrate that mSHINE outperforms other state-of-the-art HIN embedding methods.
\end{abstract}

}

\maketitle
\setcounter{page}{1}

\IEEEdisplaynontitleabstractindextext

%
\IEEEpeerreviewmaketitle

\input{Sections/Section_Introduction.tex}

%
%
%
%



\input{Sections/Section_Preliminaries.tex}
\input{Sections/Section_Related_work.tex}
\input{Sections/Section_Proposed_Method.tex}
\input{Sections/Section_Experiments.tex}

\input{Sections/Section_Conclusion.tex}
\bibliographystyle{IEEEtran}
\bibliography{IEEEabrv,HIN}
%
\begin{IEEEbiography}[{\includegraphics[width=1in,height=1.25in,clip,keepaspectratio]{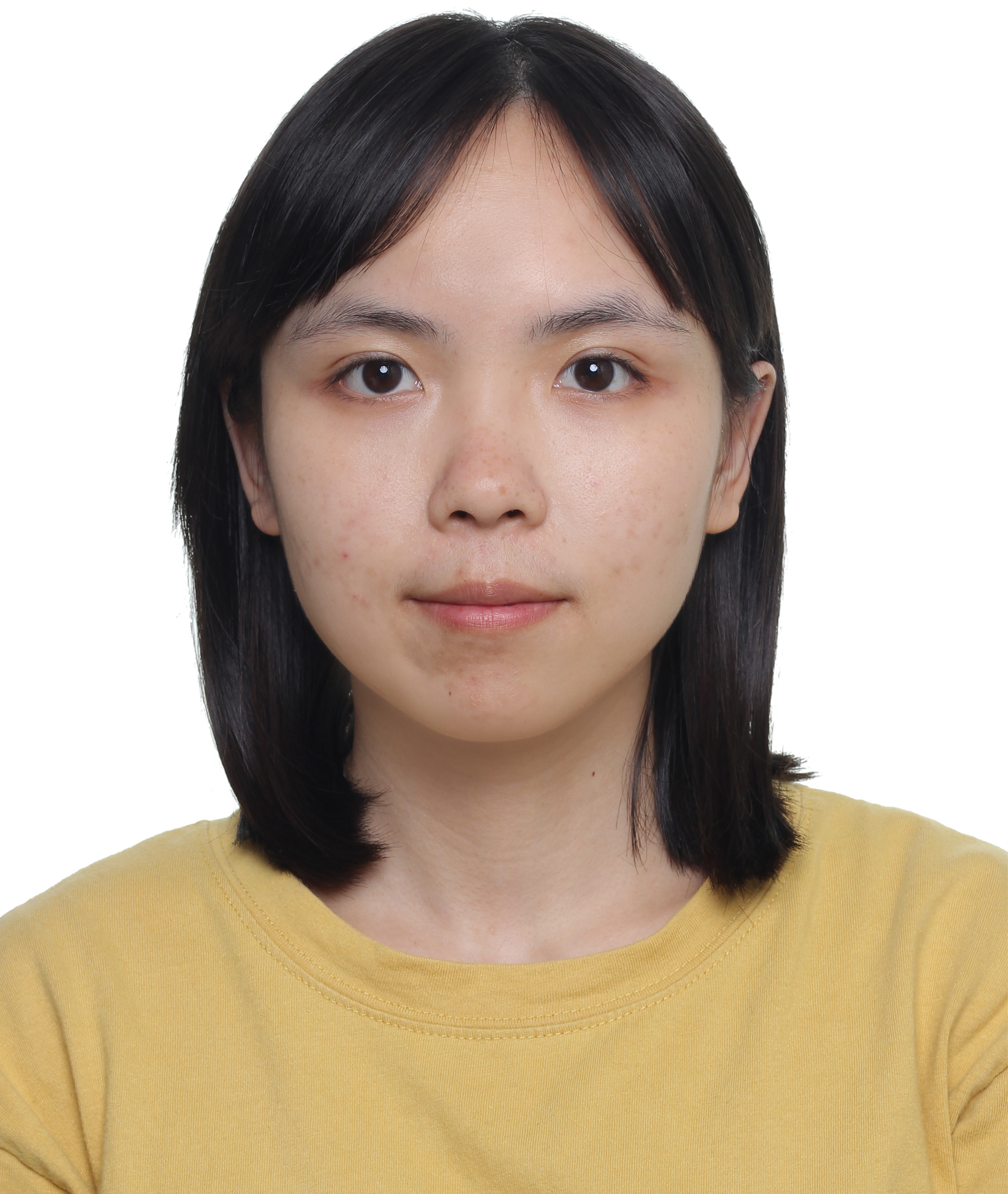}}]{Xinyi Zhang}
received the master's degree in science from the Nanyang Technological University(NTU) in 2017. She is currently working toward the PhD's degree at NTU. Her research interests including machine learning and data mining.
\end{IEEEbiography}
\begin{IEEEbiography}[{\includegraphics[width=1in,height=1.25in,clip,keepaspectratio]{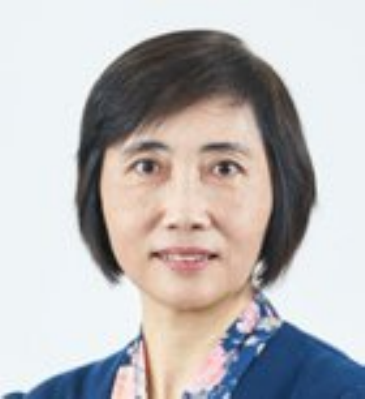}}]{Lihui Chen}
(SM’07) received the B.Eng. degree in computer science and engineering from Zhejiang University, Hangzhou, China, and the Ph.D. degree in computational science from the University of St Andrews, St Andrews, U.K. She is currently an Associate Professor with the Centre for Info-comm Technology, School of Electrical and Electronic Engineering, Nanyang Technological University, Republic of Singapore. She has authored over 110 referred papers in international journals and conferences in her research areas. Her current research interests include machine learning algorithms and applications, graph data analytics, text/data mining, data clustering.
\end{IEEEbiography}





\vfill


\end{document}

%% file: Sections/Section_Introduction.tex
    \IEEEraisesectionheading{
    \section{Introduction}
    \label{sec:Introduction}}
    \IEEEPARstart{I}{nformation} network(Graph) is a type of structural data which can be used to describe a set of objects and their relationships. This type of data is widely used in representing social networks, knowledge graphs, world wide web and so on. Therefore, in the era of information and technology, the topic of information network analysis has received a lot attention from the research community to industrial sectors. Recent works \cite{reference:perozzi2014deepwalk,reference:grover2016node2vec,reference:shi2018easing,reference:cui2019survey} have shown the effectiveness of network embedding in information network analysis. The goal of network embedding is to convert graph data into a lower dimensional space while the important information (e.g., structure information and attribute information) is still well preserved so that these lower-dimensional representations are able to facilitate downstream HIN analysis tasks such as node classification\cite{reference:cai2018graphembedding} and link prediction\cite{reference:hu2018leveraging,reference:Shi2019Herec,reference:Epasto2019splitter}. Network embedding is also referred as network representation learning in this paper. 
    
\begin{figure}[!t]
    \centering
    \includegraphics[width=2in]{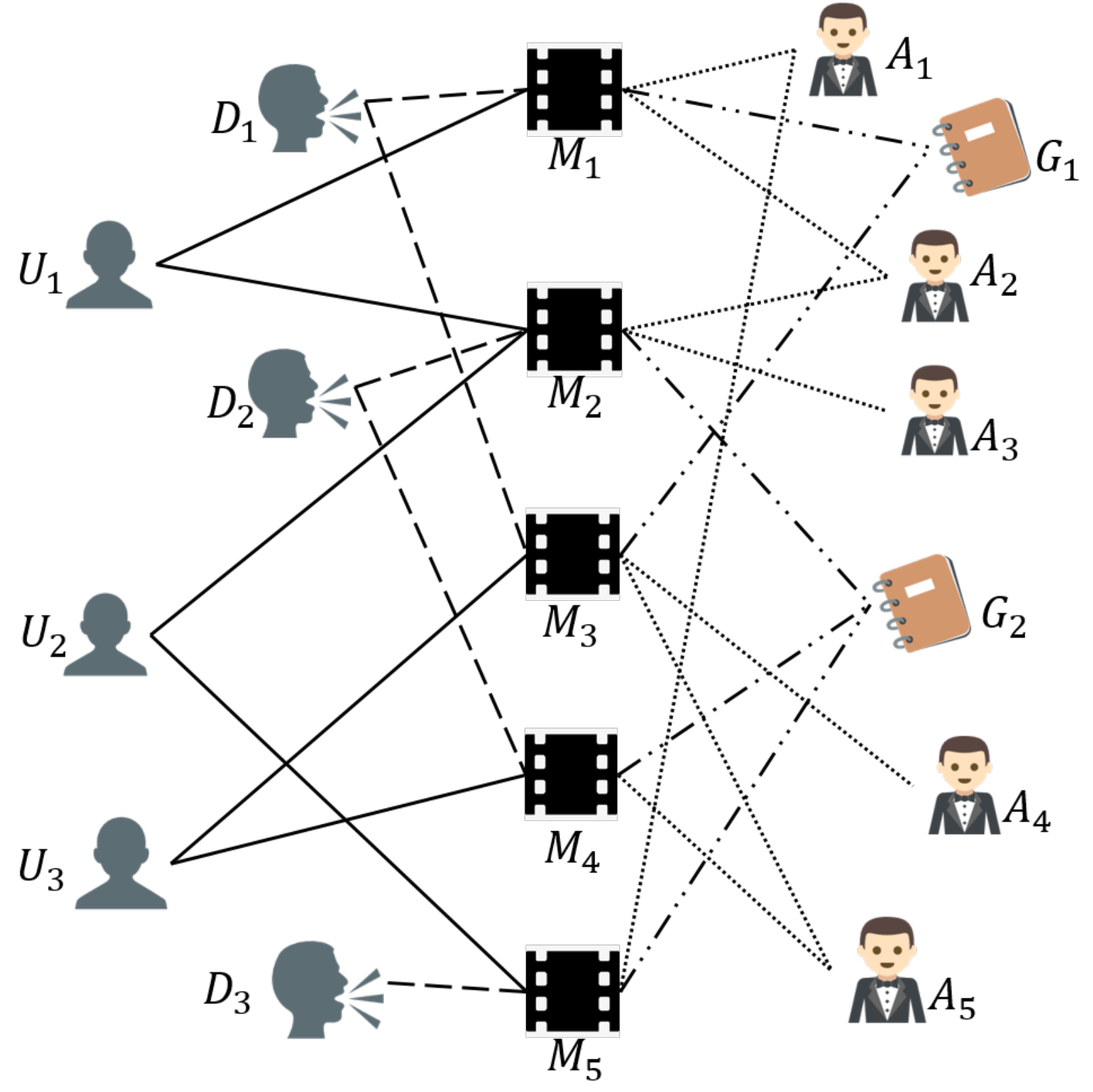}    
    \caption{An example of IMDB information network and $U_i$, $M_i$, $A_i$, $G_i$, $D_i$ denote different Users, Movies, Actors, Genres and Directors respectively.}
    \label{fig_hin}
    \end{figure}

    Recently, heterogeneous information networks (HINs), a special type of information networks which contain various types of nodes and edges, is receiving increasing attention because of its strong capability of describing objects with rich information\cite{reference:shi2016survey}. However, the heterogeneity of HINs also raises the problem of potential semantic incompatibility which might affect the performance of network embeddings in downstream tasks. For example, \figurename\ref{fig_hin} illustrates a simple HIN where users are connected with their reviewed movies, some attributes (actors, directors and genres) of these movies are also listed. If we analyse the relationship between $U_1$ and $U_3$, it is easy to find that they reviewed movies produced by the same directors. However, as for $U_1$ and $U_2$, they reviewed movies with the same actors. So we may conclude that $U_1$ and $U_2$ are interested in the same actors while $U_1$ and $U_3$ are interested in the same directors. This reflects different types of semantic information of HINs which can also be referred as different aspects of HINs. If we focus on actors, $U_1$ should be closer to $U_2$ while it should be closer to $U_3$ if we focus on directors. In this case, directly representing each user with a single lower-dimensional representation may lead to information loss and reduce the quality of network embeddings.
    
    Since traditional homogeneous information network embedding methods \cite{reference:perozzi2014deepwalk,reference:grover2016node2vec,reference:Tang2015LINE,reference:wang2018graphgan} fail to capture information included in HINs from different aspects, some new methods \cite{reference:metapath2vec,reference:tang2015pte,reference:shi2018easing,reference:fu2017hin2vec,reference:chen2017task,reference:shi2018aspem,reference:liu2019single,reference:hu2019HeGAN,reference:Shi2019Herec} that are designed for HIN embedding have been proposed to address the problems mentioned above. Many of these works make use of meta-paths\cite{reference:sun2011pathsim} to distinguish different semantic information and have verified the effectiveness of meta-paths in HIN embedding. However, meta-path selection is still a challenge in current research works. Some methods\cite{reference:metapath2vec,reference:Shi2019Herec} require selecting meta-paths in advance based on experts experience and some other methods\cite{reference:fu2017hin2vec,reference:chen2017task} require setting the limitation of meta-paths' length and then all possible meta-paths will be greedily examined. These methods are hard to be applied in practical applications when there is no prior knowledge from domain experts for optimal meta-path selection. Other tools\cite{reference:liu2019single,reference:shi2018aspem} instead of meta-paths can also be applied to represent different aspects of an HIN while the practical meaning of these aspects are sometimes unexplainable. Besides, some hyper-parameters are still required to be set to select valid aspects for practical using. Another challenge is, for each of the selected meta-path(aspect), re-learning all node representations is required and this is extremely inefficient especially when a large number of meta-paths(aspects) or meta-path(aspect) combinations need to be tested.
    
    To address the problems mentioned above, we propose \textbf{mSHINE}( a \textbf{m}ultiple-meta-paths \textbf{S}imultaneous learning framework for \textbf{H}eterogeneous \textbf{I}nformation \textbf{N}etwork \textbf{E}mbedding) which takes an HIN as input and learns multiple node representations for each node as the output, where each representation is associated with one respective meta-path. This framework mainly consists of two parts: \textit{Initial meta-path selection module} and \textit{Node representation learning module}. Firstly, to reduce the cost of meta-path exploration during training stage, a set of criteria are proposed to select valid initial meta-paths. Here, initial meta-paths refer to the meta-paths that are selected for training. Then the node representation learning module, which is inspired by the concept and basic structure of recurrent neural network(RNN)\cite{reference:elman1990finding}, is applied to capture long-term node information. At the same time, combined with meta-paths information, the representation learning module is designed to be capable of simultaneously learning the respective node representations for all selected initial meta-paths. Finally, node representations learned from different meta-paths can be applied to the target tasks. Users can further choose the most suitable meta-path which achieves the best performance for a specific task. In addition to the learned node representations, the trained framework can also be easily applied in downstream link prediction tasks which is a type of common and critical HIN analysis task. The main contributions of this work are as follows:
    
    \begin{itemize}[leftmargin=*]
        \item Proposed mSHINE, a novel HIN representation learning framework includes two key components: node representation learning module and initial meta-path selection module.
        
        \item The representation learning module is designed to facilitate \textbf{simultaneously learning} multiple node representations for all selected meta-paths. This improves the feasibility and learning efficiency when multiple meta-paths need to be considered.
        
        \item The representation learning module combines the concept of meta-path with RNN-inspired structure to \textbf{capture different semantic and long-term node information}, thereby solving the problem of incompatibility of HINs to a certain extent. 
        
        \item The initial meta-path selection module, consisting of a set of selection criteria, is proposed to\textbf{ select initial meta-paths}. This module is helpful to reduce the meta-path exploration cost in the beginning.
        
        \item The well trained mSHINE \textbf{can be easily applied in downstream HIN analysis tasks} such as link prediction and node classification. 
        
        \item The effectiveness of the proposed mSHINE framework is verified via extensive experiments study on 5 real-world datasets.
    \end{itemize}
    
    The rest of this paper is organized as follows. Section \ref{sec:Preliminaries} describes notations used in this paper and some necessary concepts which are widely used in HIN representation learning works. A critical review of the current research work that is related to HIN representation learning is given in Section \ref{sec:Related Work}. The details of the proposed representation learning module as well as the criteria we proposed to select initial meta-paths are described in Section \ref{sec:proposed_method}. A number of experiments are conducted and detailed analysis are illustrated in Section \ref{sec:experiments}. Finally, the conclusion is given in Section \ref{sec:conclusion}.

%% file: Sections/Section_Preliminaries.tex
\section{Preliminaries}
\label{sec:Preliminaries}
We define related concepts and notations in this section.

\begin{definition} {\bf Heterogeneous Information Network(HIN)}. 
A heterogeneous information network, denoted as $\mathcal{G = (V,E)}$, is defined as a directed graph which consists of a node set $\mathcal{V}$ and an edge set $\mathcal{E}$. The heterogeneous information network is associated with a node type mapping $\phi:\mathcal{V} \to T_v$ and an edge type mapping $\psi:\mathcal{E} \to T_e$. $T_v$ and $T_e$ denote the sets of node and edge types and $\left| T_v \right|+\left|T_e\right| > 2$.
\end{definition}

\begin{definition}{\bf Network Schema}. Denoted as $T_\mathcal{G}=(T_v,T_e)$, network schema is a meta template for a heterogeneous information network $\mathcal{G}$. It is a graph defined over node types $T_v$ with edges as relations from $T_e$. Network schema defines the type constraints on the sets of nodes as well as the allowed relationships between these nodes.
\end{definition}

\begin{definition} {\bf Meta-path}. A meta-path is a sequence of node and edge types between 2 given nodes which is defined on the heterogeneous network schema to explain how the nodes are related. A meta-path is denoted in the form of $T_{V_1} \xrightarrow{T_{E_1}} T_{V_2} \xrightarrow{T_{E_2}} \cdots T_{V_{l-1}} \xrightarrow{T_{E_{l-1}}} T_{V_l}$, which defines a composite relation $ T_{E_{1,2,\cdots,l-1}}=T_{E_1}\circ T_{E_2}\circ \cdots \circ T_{E_{l-1}}$ between node types $T_{V_1}$ and $T_{V_{l}}$, where $\circ$ denotes the composition operator on relations. A meta-path is referred as symmetric meta-path if the composite relation $T_{E_{1,2,\cdots,l-1}}$ defined by it is symmetric. In HIN representation learning, many previous work \cite{reference:metapath2vec}\cite{reference:sun2011pathsim}\cite{referernce:Sun2012MiningHIN} has shown the effectiveness of learning representations based on different meta-paths where each meta-path can also be considered as an aspect of HIN.
\end{definition}

\begin{definition}{\bf Meta-path-based Random Walk}. Given an HIN $\mathcal{G=(V,E)}$ and a meta-path $M_i$ : $T_{V_1} \xrightarrow{T_{E_1}} \cdots T_{V_i} \xrightarrow{T_{E_i}} T_{V_{i+1}} \cdots \xrightarrow{T_{E_{l-1}}} T_{V_l}$, the random walk is generated based on the following transition probability at the $n$-th step:
 \begin{IEEEeqnarray}[\setlength{\nulldelimiterspace}{0pt}]{rls}
        p(V_{n+1}& | V_n, M_i)=\IEEEnonumber \\
        &\frac{1}{|\mathcal{N}^{T_{V_{i+1}}}(V_n)|}, & $(V_n,V_{n+1})\in \mathcal{E}$ and $\phi(V_{n+1}) = T_{V_{i+1}}$\IEEEnonumber\IEEEnosubnumber\\*[-0.625\normalbaselineskip]
        \smash{\left\{\IEEEstrut[6\jot][6\jot]\right.}&&\IEEEnonumber\\*[0.0\normalbaselineskip]
        &0,&otherwise\IEEEnonumber\IEEEnosubnumber
\end{IEEEeqnarray}

where $\phi(V_n)=T_{V_i}$ and $\mathcal{N}^{T_{V_{i+1}}}(V_n)$ represents the neighbor nodes of $V_n$ with the node type of $T_{V_{i+1}}$. Meta-paths are usually applied in a symmetric manner\cite{reference:metapath2vec,reference:dblp_Data} which facilitates the guidance for recursive random walk so that the walk can repetitively follow the pattern of the meta-path $M_i$ until it reaches the pre-defined node sequence length.
\end{definition}

\begin{figure}[!t]
\centering
\includegraphics[width=2.2in]{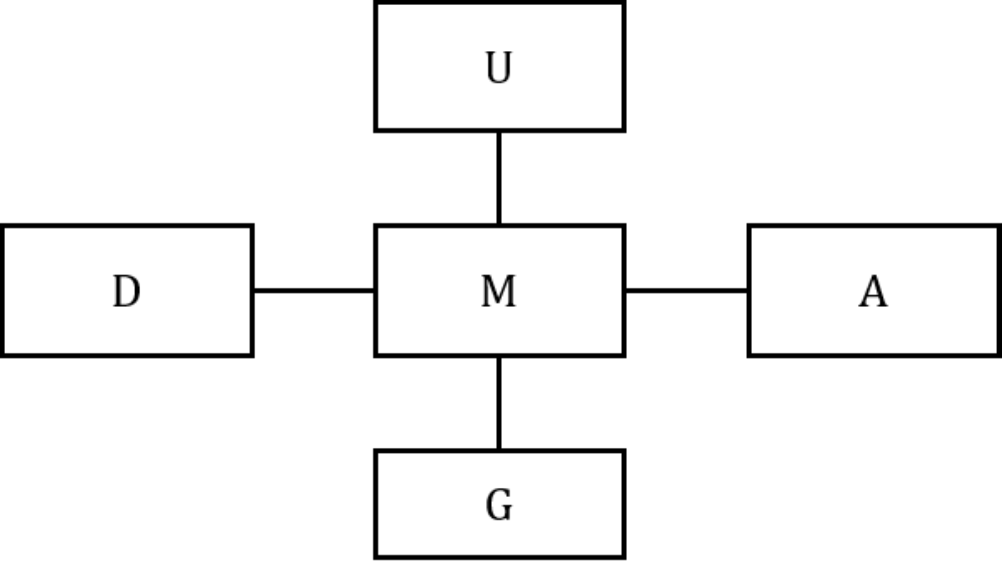}
\caption{Network schema of IMDB. Five types of nodes and four types of edges are displayed where $U$, $M$, $A$, $G$, $D$ denote the node type of User, Movie, Actor, Genre and Director respectively.}
\label{fig_hin_schema}
\end{figure}

\figurename\ref{fig_hin} illustrates a simple example of HIN with five different types of objects(User(U), Movie(M), Actor(A), Genre(G) and Director(D)) and four different types of relationships(user reviewed movie(U-M), actor played movie(A-M), director directed movie(D-M), genre of movie(G-M)). \figurename\ref{fig_hin_schema} displays the network schema of the HIN showed in \figurename\ref{fig_hin} in which nodes U, M, A, G, D represent the object type of User, Movie, Actor, Genre and Director respectively. Different types of relationships between different types of objects are also displayed. If we do random walk on this HIN based on the symmetric meta-path $U$$ \xrightarrow{}$$M$$\xrightarrow{}$$A$$\xrightarrow{}$$M$$\xrightarrow{}$$U$, the extracted node sequences can be $U_1$$ \xrightarrow{}$$M_1$$\xrightarrow{}$$A_2$$\xrightarrow{}$$M_2$$\xrightarrow{}$$U_2$$\xrightarrow{}$$M_5$$\xrightarrow{}$$A_1$$\xrightarrow{}$$M_1\cdots$.

The notations that are used in this article are summarized in Table \ref{table_notations_table} for easy reading.

\begin{table}[!t]
    \renewcommand{\arraystretch}{1.3}
    \caption{Notations and Explanations}
    \label{table_notations_table}
    \centering
        \begin{tabular}{|c||c|}
        \hline
        \bfseries Notation & \bfseries Explanation\\
        \hline
        $\mathcal{M}$       & Meta-path set\\\hline
        $M_i$               & The $i$th meta-path\\\hline
        $\mathcal{D}$       & The sampled node sequences\\\hline
        $\mathcal{D}^{M_i}$ & \makecell[c]{Node sequences which are \\extracted based on meta-path type $M_i$}\\\hline
        $V_n$/$E_n$         & Node $n$/ Edge $n$\\\hline
        $T_{V_n}$/$T_{E_n}$ & Node type of node $V_n$ /Edge type of edge $E_n$\\\hline
        $T_{E_{ab}}$        & \makecell[c]{Node sequence type of a three-element\\ node sequence $V_{n-1} \xrightarrow{T_{E_a}} V_{n} \xrightarrow{T_{E_b}} V_{n+1}$} \\\hline
        $\Vec{r}_{T_{E_{ab}}}$  &   \makecell[c]{Relation representation of a three-element\\ node sequence with the type of $T_{E_{ab}}$}\\\hline
        $\Vec{x}_{V_n}$/$\Vec{h}_{V_n}$     & \makecell[c]{General basic/state node\\ representation of node $V_n$}\\\hline 
        
        $\Vec{x}_{V_n}^{M_i}$/$\Vec{h}_{V_n}^{M_i}$ & \makecell[c]{Basic/state node representation\\ of node $V_n$ in meta-path $M_i$}\\\hline
        
        $f_x^{M_i}/f_h^{M_i}/f_y^{M_i}$     & \makecell[c]{Meta-path-related functions for general\\ basic/state/target node representations}\\\hline
        $\mathcal{C}_{M_i}$ & \makecell[c]{The decomposed three-element \\node sequence type set of meta-path $M_i$}\\\hline
        \end{tabular}
\end{table}

%% file: Sections/Section_Related_work.tex

\section{Related Work}
\label{sec:Related Work}

Related studies will be reviewed in this section from two aspects: \textit{network embedding} and \textit{heterogeneous information network}.

Network embedding means to convert network data into a lower dimensional space where the important information of the network is still well preserved\cite{reference:cai2018graphembedding, reference:cui2019survey}. Some research work has shown the effectiveness of network embedding in many network analysis tasks such as node classification and clustering\cite{reference:Wang2019HAN,reference:shi2016survey}, link prediction\cite{reference:shi2018easing} and recommendation\cite{reference:Shi2019Herec,reference:fan2019MEIRec}. In DeepWalk\cite{reference:perozzi2014deepwalk}, skip-gram and negative-sampling\cite{reference:Mikolov2013Word2vec} are applied to learn node representations based on node sequences sampled through random walk. It is also the first representation learning method which combines language modelling and random walk to explore information networks. Furthermore, Node2vec\cite{reference:grover2016node2vec} proposes to conduct a more flexible exploration strategy when sampling node sequences. Unlike DeepWalk and Node2vec where no clear objective about what network properties are preserved, LINE\cite{reference:Tang2015LINE} is designed to preserve both of the first-order and the second-order proximities in networks. To improve the effectiveness of learned node representations, some algorithms\cite{reference:Cheng2016TAWD,reference:tang2015pte} which are designed for special network types(e.g., networks whose nodes represent different papers and the paper content is known.) are proposed to integrate richer node information. Some neural network based works\cite{reference:wang2016SDNE,reference:zhang2018anrl} apply auto-encoder to information network to do dimensionality reduction. However, these neural network based auto-encoders have difficulties in handling large sparse networks which might lead to expensive computation cost and sub-optimal performance. Then, GNNs\cite{reference:kipf2016semi,reference:hamilton2017inductive,reference:velickovic2018GAT} are proposed to iteratively aggregate neighbor node information and update center node embedding through activation function which also show impressive performance in several tasks. Because of the concept of receptive-field used in GNNs, these GNNs are referred as graph convolutional neural network methods. However, compared with random-walk-based methods\cite{reference:perozzi2014deepwalk,reference:grover2016node2vec}, GNN-based methods usually fail to capture long-term relationships between nodes which is usually critical in node classification tasks. In addition, to achieve satisfied performance, GNN-based methods usually require suitable initialization which makes GNN-based methods harder to be applied in some applications. Furthermore, some recurrent neural network based methods\cite{reference:su2019rnngraph,reference:li2015gated} are also proposed to process networks. The information propagation in GGS-NNs\cite{reference:li2015gated} between nodes follows the structure of RNN and in theory, there is no limitation of the length of information propagation. All methods introduced above are designed for homogeneous information network and they fail to handle the problem of incompatibility in HINs.

Heterogeneous information networks(HINs)\cite{reference:shi2016survey} currently are widely used to model complex data with different types of objects and relationships. However, traditional network embedding methods introduced above can hardly handle complex information provided by HINs. Hence, some tools are designed to separately describe different semantic information of HINs. For example, meta-path\cite{reference:sun2011pathsim}, a tool that is widely used to distinguish different semantic information in HIN has been verified its effectiveness. To be more specific, Metapath2vec\cite{reference:metapath2vec} proposed to prepare training node sequence based on different meta-paths and separately apply skip-gram to learn node representations for each meta-path. Besides, they also proposed heterogeneous negative-sampling which constrains the node type of negative samples to be the same as the target node and this technique becomes popular in later methods. However, Metapath2vec requires manually selecting meta-paths based on experts' experience so it is hard to be applied when no prior knowledge is available. Besides, re-training node representations for different meta-paths is extremely inefficient when the number of possible meta-paths is huge. To address the difficulty of re-training node representations, HAN\cite{reference:Wang2019HAN} proposed to simultaneously extract node features under different meta-paths with multiple GATs\cite{reference:velickovic2018GAT} and then use attention mechanism\cite{reference:yang2016hierarchical} to combine these node features while how to select initial meta-paths for training is not discussed. Besides, to learn attention value for meta-paths, node labels are required in HAN. To address the difficulty of meta-path selection, some works\cite{reference:fu2017hin2vec,reference:chen2017task} are proposed to greedily analyse all possible meta-paths within the default meta-path length limitation. However, this length limitation still needs presetting. In addition to meta-paths, some methods\cite{reference:liu2019single,reference:shi2018aspem} also proposed to recognise different aspects of HINs by directly analysing the structure information of HINs while the practical meaning of these detected aspects are usually unexplainable. HeGAN\cite{reference:hu2019HeGAN} and HEER\cite{reference:shi2018easing} directly consider different edge types as different aspects which is equivalent to omitting the procedure of aspects exploration. This is sub-optimal in exploring different semantic information of HINs. In addition, in order to achieve satisfied performance, HeGAN also requires initializing carefully. For example, in the original paper of HeGAN, the node representations learned from Metapath2vec are used to initialize HeGAN, which is extremely inefficient in practical applications.


%% file: Sections/Section_Proposed_Method.tex


\section{mSHINE}
\label{sec:proposed_method}
    Given an HIN $\mathcal{G = (V,E)}$, mSHINE intends to find a set of initial meta-paths $\mathcal{M}=\{M_1, M_2, \dots, M_c\}$(unless initial meta-paths set is given based on prior knowledge) and then concurrently learn a total of $c$ $d$-dimensional representations for each node $V_i \in \mathcal{V}$. The important information of $\mathcal{G}$ should be well preserved in the learned node representations. The trained module can be easily applied in downstream link prediction tasks and achieve state-of-the-art performance.
    
\begin{figure*}[tb]
        \centering
        \includegraphics[width=5.8in]{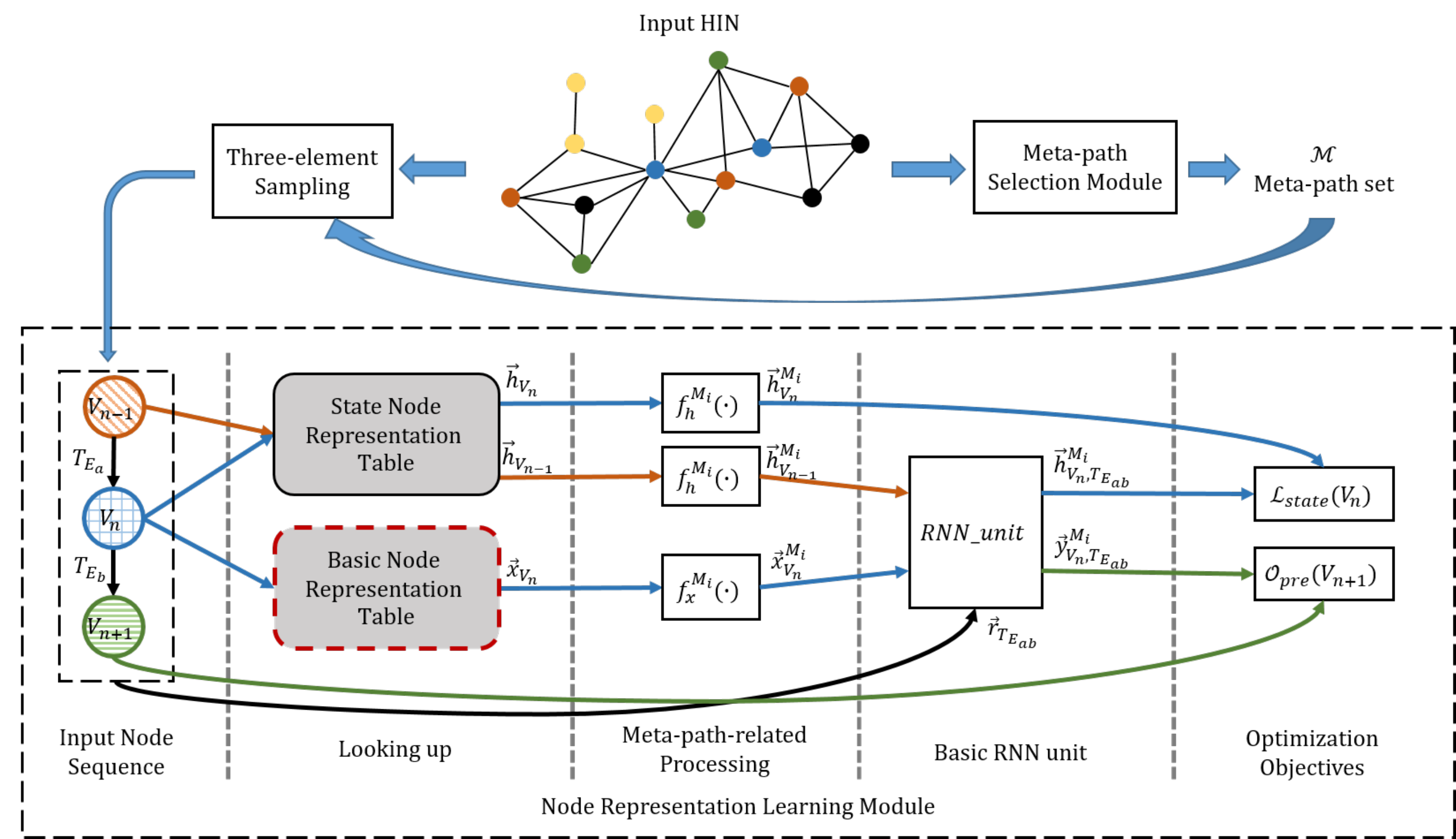}
        \caption{The schematic illustration of mSHINE. Meta-path selection criteria are applied to select valid meta-paths first and these valid meta-paths are used to direct sampling three-element node sequences. For each three-element node sequence $V_{n-1}$$\xrightarrow{T_{E_a}}$$V_{n}$$\xrightarrow{T_{E_b}}$$V_{n+1}$, the general basic node representation $\Vec{x}_{V_{n}}$ for $V_{n}$ and general state node representations $\Vec{h}_{V_{n-1}}$, $\Vec{h}_{V_n}$ for $V_{n-1}$ and $V_n$ are generated through looking up the basic and state node representation tables. Then meta-path-related functions are applied to process the general basic and state node representations to generate $\Vec{x}_{V_n}^{M_i}$, $\Vec{h}_{V_{n-1}}^{M_i}$ and $\Vec{h}_{V_n}^{M_i}$ respectively. The processed node representations($\Vec{x}_{V_n}^{M_i}$ and $\Vec{h}_{V_{n-1}}^{M_i}$) are then fed into RNN-inspired unit together with the edge representation $\Vec{r}_{T_{E_{ab}}}$ to generate a new state node representation $\Vec{h}_{V_{n},T_{E_{ab}}}^{M_i}$ for $V_n$ as well as the prediction of the next node $V_{n+1}$. All the trainable parameters in the proposed structure are trained based on $\mathcal{L}_{state}(V_{n})$ and $\mathcal{O}(V_{n+1})$. Finally, the basic node representations as well as meta-path-related functions can be used to generate meta-path-related node representations and the trained node representation learning module can be applied in downstream link prediction tasks.}
        \label{fig_representation learning}
        \end{figure*}

    The overall structure of mSHINE is shown in \figurename\ref{fig_representation learning} which mainly consists of two modules and this section will introduce these modules in detail respectively:
    \begin{inparaenum}[1)]
    \item \textbf{Node representation learning} module and
    \item \textbf{Initial meta-path selection} module.
    \end{inparaenum}
    
    \subsection{Node Representation Learning}
    \label{subsect:Rpresentation Learning}
        As introduced in Section \ref{sec:Introduction} and Section \ref{sec:Related Work}, traditional network embedding methods such as DeepWalk \cite{reference:perozzi2014deepwalk} and Node2vec \cite{reference:grover2016node2vec} fail to capture different semantic information of HINs since they simply use random walk to generate node sequences for node representation learning. Hence, we choose to extract meaningful node sequences under the direction of meta-paths for node representation learning to distinguish different semantic information of an HIN. This is similar to Metapath2vec \cite{reference:metapath2vec} where an HIN is converted to a stream of node sequences through meta-path-based random walk. By considering these extracted node sequences as sentences and nodes as words, language modeling methods can be applied to explore the property of the original HIN. Different from Metapath2vec\cite{reference:metapath2vec} where Word2vec\cite{reference:Mikolov2013Word2vec} is chosen as the language modeling method, the node representation learning module in mSHINE is developed by adapting the working principle and basic structure of RNN which is designed for processing sequential data. The detailed structure of this representation learning module will be introduced from two levels:
        \begin{inparaenum}[1)]
        \item Structure Information Modelling which is designed to capture connection information from extracted node sequences.
        and \item Semantic Information Modelling which is developed on the basis of structure information modelling to include semantic information.
        \end{inparaenum}
        \begin{figure}[t]
        \centering
        \includegraphics[width=1.4in]{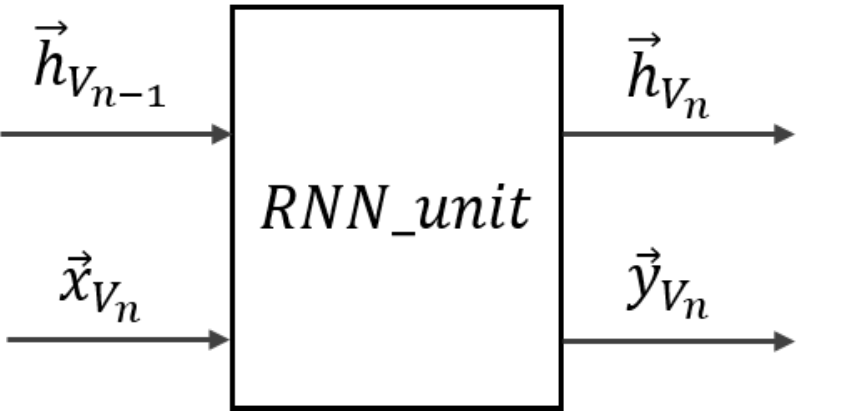}
        \caption{Basic RNN unit for node sequence processing.}
        \label{fig_basic_rnn_unit}
        \end{figure}
        \subsubsection{Structure Information Modelling}
        \label{subsubsec:Structure Information Modelling}
        At the structure information modelling level, the edge and node types are all ignored since we only focus on connection information at this level.  
        
        In the original RNN structure, at each timestamp $t$, an RNN unit takes the current \textit{input} $\Vec{x}_t$ and the \textit{hidden state} $\Vec{h}_{t-1}$ to predict the current \textit{output} $\Vec{y}_t$ and generate a new \textit{hidden state} $\Vec{h}_{t}$. In this work, when we process an extracted node sequence(the details about how to select initial meta-path and how to extract meta-path-based node sequences is explained later) with RNN structure, the node sequence should be decomposed into multiple short three-element sequences and each three-element sequence corresponds a timestamp. For example, supposing the node sequence is 
        $V_0$$\xrightarrow{}$$V_1$$\cdots$$V_{l-1}$$\xrightarrow{}$$V_{l}$ 
        and it can be decomposed into a three-element sequences set 
        $\{$$V_0$$\xrightarrow{}$$V_1$$\xrightarrow{} $$V_{2}$,$V_1$$\xrightarrow{}$$V_2$$\xrightarrow{} $$V_{3}$,$\cdots$,$V_{l-2}$$\xrightarrow{} $$V_{l-1}$$\xrightarrow{}$$V_{l}$$\}$, for each three-element sequence in this set, the processing procedure is shown in \figurename \ref{fig_basic_rnn_unit} and formally, it can be written as:
        \begin{equation}
            \label{equ:ori_1}
            \Vec{h}_{V_{n}} = \sigma(W_{xh}\Vec{x}_{V_n} + W_{hh}\Vec{h}_{V_{n-1}})
        \end{equation}
        \begin{equation}
            \Vec{y}_{V_{n}} = W_{hy}\Vec{h}_{V_{n}}
        \end{equation}
        \begin{equation}
        \label{equ:ori_3}
            p(V_{n+1}|V_{n} \cdots V_{1}) = \frac{\exp{(\Vec{y}_{V_{n}(V_{n+1})})}}{\sum_{v' \in \mathcal{V}}\exp{(\Vec{y}_{V_{n}(v')})}}
        \end{equation}
        The input of the RNN unit should be the basic representation of $V_n$($\Vec{x}_{V_n}$) and the state representation of $V_{n-1}$($\Vec{h}_{V_{n-1}}$). The output includes a new state representation $\Vec{h}_{V_n}$ for $V_n$ as well as the probability of the existing of $V_{n+1}$ which is denoted as $p(V_{n+1} |V_n \cdots V_1)$ and it is calculated from $\Vec{y}_{V_n}$. $W_{xh}\in R^{d \times d}$ and $W_{hh} \in R^{d \times d}$ denote the trainable transform matrices which are used to do transformation between the basic representation domain and the state representation domain. $d$ represents the dimension of node representations. $\Vec{y}_{V_n(v')}$ denotes the possibility of the existing of node $v'$ based on the prediction of node sequence $V_n \cdots V_1$. $W_{hy} \in R^{N \times d}$ denotes the output representation matrix which contains all the target node representations and $N$ here represents the number of nodes in this HIN. $\sigma(\cdot)$ is the activation function.
        
        
        So far, by optimizing the probability of the existing of $V_{n+1}$, the structural information of the extracted node sequences can be captured by the basic RNN-inspired structure and preserved in the learned basic and state node representations. 
        
        
        
        
        
        \subsubsection{Semantic Information Modelling}
        In HINs, besides structural information, semantic information which is provided by the attributes of different edges and nodes also plays an important role in describing object properties. Based on the basic representation learning module introduced in Section \ref{subsubsec:Structure Information Modelling}, we will include more semantic information at this level to enhance the quality of the learned node representations. This level mainly consists of two parts: 
        \begin{inparaenum}[1)]
            \item Meta-path-based semantic information and
            \item Edge-based semantic information.
        \end{inparaenum}
         
         \textbf{Meta-path-based semantic information}. As introduced in Section \ref{sec:Preliminaries}, node representations which are learned based on different meta-paths can reflect different aspects of this node and separately embedding these aspects is helpful to describe different properties of this node. So, when processing a node sequence which is extracted based on the meta-path $M_i$, the corresponding meta-path information should be embedded into the representation of each node in this sequence. This is achieved through the meta-path-related functions and the procedure can be formally written as:
        \begin{equation}
             \Vec{x}_{V_n}^{M_i} = f_{x}^{M_i}(\Vec{x}_{V_n})
        \end{equation}
        \begin{equation}
        \label{equ:metapath_related_hidden}
             \Vec{h}_{V_n}^{M_i} = f_{h}^{M_i}(\Vec{h}_{V_n})
        \end{equation}
        Here, $\Vec{x}_{V_n}$ and $\Vec{h}_{V_n}$ denote the general basic node representation and general state node representation of $V_n$ respectively. The word \textit{general} here indicates that these representations contain node information from all meta-paths and they are independent from any specific meta-paths. Then meta-path-related functions $f_x^{M_i}(\cdot)$ and $f_h^{M_i}(\cdot)$ are used to decode information that is related to the meta-path $M_i$ from $\Vec{x}_{V_n}$ and $\Vec{h}_{V_n}$ respectively. As a result,  $\Vec{x}_{V_n}^{M_i}$ and $ \Vec{h}_{V_n}^{M_i}$ can be used to reflect the meta-path-related properties of $V_n$ and these meta-path-related node representations can be easily used in downstream HIN analysis tasks.
        
        \textbf{Edge-based semantic information}. As introduced in Section \ref{subsubsec:Structure Information Modelling}, a node sequence which is extracted based on the meta-path $M_i$ can be further decomposed into multiple three-element sequences ($V_{n-1}$$\xrightarrow{T_{E_a}}$$V_{n}$$ \xrightarrow{T_{E_b}}$$V_{n+1}$). The three-element node sequence type $T_{E_{ab}}$ (a.k.a edge-based information) is also a part of the crucial information which needs to be embedded into the representation of each node that appears in this three-element sequence. To include both edge-based and meta-path-based information into the node representations, the procedure which is originally written as Eq.\eqref{equ:ori_1}-\eqref{equ:ori_3} can be further modified as:
        \begin{equation}
        \label{equ:gen_hidden}
            \Vec{h}_{V_{n},T_{E_{ab}}}^{M_i} = \sigma(W_{xh}\Vec{x}_{V_n}^{M_i} + W_{hh}\Vec{h}_{V_{n-1}}^{M_i} + W_{rh}\Vec{r}_{T_{E_{ab}}})
        \end{equation}
        \begin{equation}
            \Vec{y}_{V_{n},T_{E_{ab}}}^{M_i} = W_{hy}^{M_i} \Vec{h}_{V_{n},T_{E_{ab}}}^{M_i}
        \end{equation}
        where
        \begin{equation}
            W_{hy}^{M_i} = f_y^{M_i}(W_{hy})
        \end{equation}{}
        and
        \begin{IEEEeqnarray}{Cl}
            p(V_{n+1}|V_{n} \cdots &V_{1},M_i,T_{E_{ab}})=\IEEEnonumber \\
            &\frac{\exp{(\Vec{y}_{V_{n},T_{E_{ab}}(V_{n+1})}^{M_i})}}
            {\sum\limits_{v' \in \mathcal{V}, \phi(v')\in \phi(V_{n+1})}\exp{(\Vec{y}_{V_{n},T_{E_{ab}}(v')}^{M_i})}} \IEEEyesnumber \label{equ:probability}
        \end{IEEEeqnarray}
        where $\Vec{r}_{T_{E_{ab}}} \in R^{d \times 1}$ is used to decode the edge-based semantic information from node representations and it is decided by the three-element node sequence type $T_{E_{ab}}$. The role of $\Vec{r}_{T_{E_{ab}}}$ can be considered as the same as the context vector $\Vec{c}$ in RNN Encoder–Decoder work\cite{reference:cho2014RNNEncoder}. Since the three-element node sequence type $T_{E_{ab}}$ is uniquely determined by the node sequence $V_{n} \cdots V_{1}$, $p(V_{n+1}|V_{n} \cdots V_{1},M_i,T_{E_{ab}})$ can be directly written as $p(V_{n+1}|V_{n} \cdots V_{1},M_i)$ in the following discussion. It is worth noting that $W_{hy}$ is row-wisely processed with its corresponding meta-path-related function $f_y^{M_i}(\cdot)$ as well. This is because each row of $W_{hy}$ is the target representation of a node and it is more reasonable to apply meta-path-related function to these target node representations too.
        
        By including both meta-path-based and edge-based semantic information, the structure introduced above is capable of capturing both connection information as well as heterogeneous semantic information of extracted node sequences.
        
        \subsubsection{Optimization Objective}
        \label{subsubset:Optimization Objective}
        Multiple optimization objectives are set to learn the trainable parameters as well as node representations in the proposed module.
        
        The first optimization objective is the prediction objective. In specific, given a node sequence $s^{n+1}=(V_{n+1}, V_n, \cdots V_0)$ which is extracted based on meta-path $M_i$, the prediction optimization objective is to maximize the log likelihood of $V_{n+1}$ given its previous node sequence $s^{n}=(V_n, V_{n-1}, \cdots V_0)$:
        \begin{IEEEeqnarray}{Cl}
            O_{pre}(V_{n+1}, M_i) & =\IEEEnonumber \\
            &\sum_{s \in \mathcal{D}^{M_i}_{v' \xrightarrow{} V_{n}}, v' \in \mathcal{V}} \log p(V_{n+1}|s^{n},M_i) \IEEEyesnumber \label{equ:objective_1}
        \end{IEEEeqnarray}
        where $\mathcal{D}^{M_i}_{u \xrightarrow{} v}$ denotes node sequences which are extracted based on meta-path type $M_i$ and these sequences should start from node $u$ and end at node $v$. Here, the probability $p(V_{n+1}|s^{n},M_i)$ is defined as Eq. \eqref{equ:probability}. Since it is not necessary for the length of $s$ to be $n$, $p(V_{n+1}|s^{n},M_i)$ is directly written as $p(V_{n+1}|s,M_i)$ in the following part. As directly optimizing Eq. \eqref{equ:objective_1} is computationally expensive, negative sampling \cite{reference:mikolov2013distributed} is used here to improve training efficiency. Hence, Eq. \eqref{equ:probability} can be approximated as:
        \begin{IEEEeqnarray}{rl}
            \log p(&V_{n+1}|s,M_i) \approx \log \sigma(W_{hy(V_{n+1})}^{M_i} \Vec{h}_{V_{n},T_{E_{ab}}}^{M_i})\vspace{\jot}\IEEEnonumber \\
            &+
            \frac{1}{K}\sum\limits_{i=1,\phi(V_{i})= \phi(V_{n+1})}^{K}{
                \log \sigma(-{W_{hy(V_{i})}^{M_i} \Vec{h}_{V_{n},T_{E_{ab}}}^{M_i})}} \IEEEyesnumber\label{equ:optimization_neg_sampling}
        \end{IEEEeqnarray}
        where $K$ denotes the number of negative samples and $W_{hy(V_{i})}^{M_i}$ denotes the target node representation of negative node $V_{i}$ under meta-path type $M_i$.
        
        As introduced above, prior to learning node representations with the proposed structure, a large number of node sequences ($\mathcal{D}$) should be prepared in advance by conducting meta-path-based random walk. Then for each node sequence $s$, an optimization algorithm is applied to maximize the objective value which is calculated based on Eq. \eqref{equ:optimization_neg_sampling}. However, as the node sequence length grows, training parameters based on the objective function becomes inefficient. So we proposed to relax the procedure of calculating state representation of each node so that processing a long node sequence to generate state representation for each node is not required anymore. To be specific, given a three-element sequence $V_{n-1}$$\xrightarrow{T_{E_a}}$$V_{n}$$ \xrightarrow{T_{E_b}}$$V_{n+1}$ and its corresponding meta-path $M_i$, it is not required to calculate $\Vec{h}^{M_i}_{V_{n-1}}$ based on its previous three-element node sequences, instead, $\Vec{h}^{M_i}_{V_{n-1}}$ can be directly obtained by looking up the stored general state representations which is then processed by a meta-path-related function $f^{M_i}_h(\cdot)$. In this case, the structure used in the node representation learning module becomes quite different from the original RNN and we call the structure as RNN-inspired structure simply because it has state node representations. In HIN representation learning, when predicting $V_{n+1}$ based on $V_{n-1}$ and $V_n$, the state representation of $V_{n-1}$ should actually provide its spacial environment information instead of just the sequential information generated based on only one single node sequence. So using stored general state representations which contain rich information captured from various node sequences is more desirable in this case. As a result, in addition to train the stored general state representations with prediction objective value, we also set the second optimization objective to embed newly generated information into stored state representations: minimizing the distance between the meta-path-related state representation $\Vec{h}^{M_i}_{V_{n}}$ which is generated through looking up stored representations and the newly generated meta-path-related state representation $\Vec{h}_{V_{n},T_{E_{ab}}}^{M_i}$ which is calculated through Eq. \eqref{equ:gen_hidden}. This distance can be calculated through:
        \begin{equation}
            \mathcal{L}_{state}(V_{n}, M_i) = \parallel {\Vec{h}_{V_n}^{M_i}}-\Vec{h}_{V_{n},T_{E_{ab}}}^{M_i}\parallel_{L_2}
        \end{equation}
        Storing general state representations of nodes makes sampling a huge number of long node sequences for training unnecessary, so when preparing training data, we only need to sample \textbf{three-element node sequences} around each node and find the corresponding meta-path type for each of the three-element node sequences. 
        
        Overall, the optimization objective of the proposed method can be formally written as:
        \begin{equation}
        \label{equ:Objective_function}
            \mathcal{O}_{all} = \sum_{M_i \in \mathcal{M}}[\sum_{V_n \in \mathcal{V}} \mathcal{O}_{pre}(V_{n},M_i)-\sum_{V_n \in \mathcal{V}}\mathcal{L}_{state}(V_{n},M_i)] 
        \end{equation}

        \subsection{Initial Meta-path Selection}
        \label{subsect:Metapath preparation}
        
        Initial meta-paths need to be selected in advance for training in this work. It is worth noting that initial meta-paths can be selected based on prior knowledge from domain experts if it is available. However, in most of the case, prior knowledge is unavailable in HIN representation learning. So in this section, the selection of valid initial meta-paths without prior knowledge is introduced. 
        
        \begin{figure}[!t]
        \centering
        \includegraphics[width=3.1in]{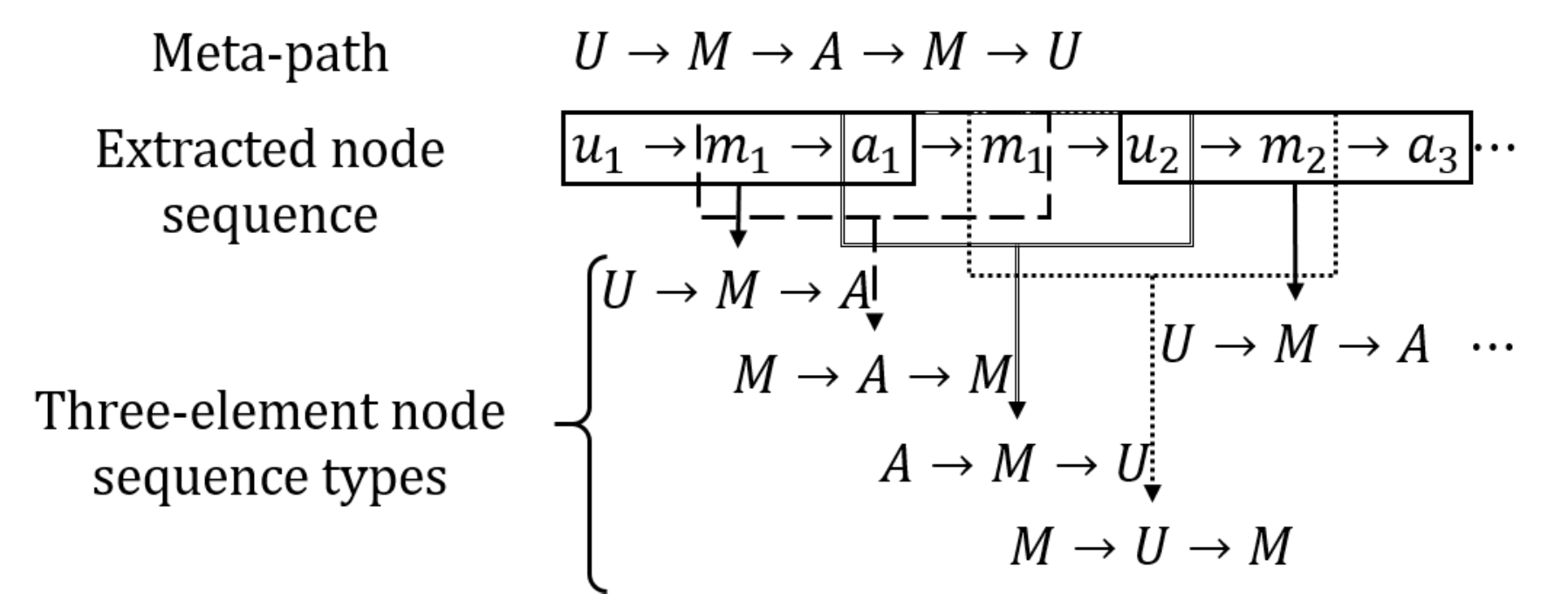}
        \caption{An example of decomposing a meta-path into the corresponding three-element node sequence type set. Based on a meta-path $M_i =$ $U$$\xrightarrow{}$$M$$\xrightarrow{}$$A$$\xrightarrow{}$$M$$\xrightarrow{}$$U$, a long node sequence can be extracted: $u_1$$\xrightarrow{}$$m_1$$\xrightarrow{}$$a_1$$\xrightarrow{}$$m_1$$\xrightarrow{}$$u_2$$\xrightarrow{}$$m_2$$\xrightarrow{}$$a_3$$\xrightarrow{}$$m_3$$\xrightarrow{}$$u_3\cdots$. Overall, there are 4 three-element node sequence types that appeared in this node sequence which are 
        $U$$\xrightarrow{}$$M$$\xrightarrow{}$$A$, $M$$\xrightarrow{}$$A$$\xrightarrow{}$$M$, $A$$\xrightarrow{}$$M$$\xrightarrow{}$$U$ and $M$$\xrightarrow{}$$U$$\xrightarrow{}$$M$. 
        So, the meta-path $M_i$ can be decomposed into a three-element node sequence type set $\mathcal{C}_{M_i} = \{$ $U$$\xrightarrow{}$$M$$\xrightarrow{}$$A$, $M$$\xrightarrow{}$$A$$\xrightarrow{}$$M$, $A$$\xrightarrow{}$$M$$\xrightarrow{}$$U$, $M$$\xrightarrow{}$$U$$\xrightarrow{}$$M$.\}. }
        \label{fig_meta-path_decompose}
        \end{figure}
        
        As discussed in Section \ref{subsubset:Optimization Objective}, the training of node representations for a meta-path $M_i$ is actually associated with a set of three-element node sequence types $\mathcal{C}_{M_i}$ and the learned node representations will be the same for meta-path $M_i$ and meta-path $M_j$ if their decomposed three-element node sequence type set $\mathcal{C}_{M_i}$ and $\mathcal{C}_{M_j}$ are the same. So two meta-paths are considered as the same if their decomposed three-element node sequence type sets are the same and we only keep the shorter meta-path during initial meta-path selection. The procedure of analysing the three-element node sequence type set $\mathcal{C}_{M_i}$ from a meta-path $M_i$ is illustrated in \figurename \ref{fig_meta-path_decompose}. Besides, according to previous work\cite{reference:sun2011pathsim,reference:Shi2019Herec}, symmetric meta-paths are more suitable for describing the similarity between two nodes with the same node types. So, we only select symmetric meta-paths as valid initial meta-paths in this paper to illustrate the idea of this framework although the proposed representation learning module also works for asymmetric meta-paths.
        
        Based on the discussion above, the criteria for selecting initial meta-paths for training are summarized as below:
        \begin{enumerate}[leftmargin=*]
            \item Only symmetric meta-paths are selected as valid meta-paths.
            \item Two meta-paths are considered as the same if their decomposed three-element node sequence type sets are the same and the shorter one is selected as valid meta-path.
            \item Any of the selected meta-paths should not be included by any other valid meta-paths.
        \end{enumerate}
        In an HIN, the number of possible three-element node sequence types is limited. So, the number of meta-paths that meet the proposed criteria is also limited and we will select all meta-paths that satisfy the criteria as initial meta-paths in mSHINE\footnote{The code to find all initial meta-paths can be found at \url{https://github.com/XinyiZ001/mSHINE}}. The node representation module introduced in Section \ref{subsect:Rpresentation Learning} is then applied to learn node representations for all initial meta-paths simultaneously. After training, node representations corresponding to different meta-paths can be easily obtained by processing general node representations with different meta-path-related functions, then we can select the most suitable meta-paths and apply the respective node representations based on specific tasks. The overall framework of mSHINE which consists of initial meta-paths selection criteria and node representation learning module is specified Algorithm \ref{Alg:proposed_method}.
        
        
        \begin{algorithm}
         \caption{mSHINE}
         \label{Alg:proposed_method}
         \algblock[Name]{Start}{End}
         \begin{algorithmic}[1]
            \Statex \textbf{Input}: 
            \Statex $\mathcal{G}$: Information network.
            \Statex \textbf{Output:}
            \Statex $\mathbf{X}$: Basic node representations.
            \Statex $f_x^M$: Meta-path-related functions.
            \Procedure {LearningModule}{$\mathcal{G}$}
            \If{Prior knowledge is available}
                \State Construct valid initial meta-path set $\mathcal{M}$  based on prior knowledge;
            \Else
                \State Construct valid initial meta-path set  $\mathcal{M}$ based on the proposed criteria in Section \ref{subsect:Metapath preparation};
            \EndIf
            \State Initialize $\mathbf{X}$, $\mathbf{H}$ and other trainable parameters;
            \While {not converged}
                \For {$M_i \in \mathcal{M}$}
                \Comment{Select a meta-path type.}
                    \State Obtain $\mathcal{C}_{M_i}$ based on $M_i$;
                    \For {$T_{E_{ab}} \in \mathcal{C}_{M_i}$}
                    \Comment{Select a three-element node sequence type.}
                        \State Sample a training sequence $s \in \mathcal{D}^{M_i}$ ($V_{n-1} \xrightarrow{T_{E_a}} V_{n} \xrightarrow{T_{E_b}} V_{n+1}$).
                        \State Sample negative samples $\Tilde{V}_{n+1}$;
                        \Start 
                            \Comment{Meta-path-related processing.} 
                            \State $\Vec{h}_{V_{n-1}}^{M_i} \leftarrow f_h^{M_i}(\Vec{h}_{V_{n-1}})$;
                            \State $\Vec{x}_{V_{n}}^{M_i} \leftarrow f_x^{M_i}(\Vec{x}_{V_{n}})$;
                            \State $\Vec{h}_{V_{n}}^{M_i} \leftarrow f_h^{M_i}(\Vec{h}_{V_{n}})$;
                            \State $W_{xy(V_{n+1})}^{M_i} \leftarrow f_y^{M_i}(W_{xy({V_{n+1}})})$;
                            \State $W_{xy(\Tilde{V}_{n+1})}^{M_i} \leftarrow f_y^{M_i}(W_{xy({\Tilde{V}_{n+1}})})$;
                        \End
                        \State Calculate objective value with Eq.\eqref{equ:Objective_function};
                        \State Update trainable parameters; 
                    \EndFor
                \EndFor
            \EndWhile
            \State \textbf{return} $\mathbf{X}$,$f_x^M$;
            \EndProcedure
        \end{algorithmic}
        \end{algorithm}

%% file: Sections/Section_Experiments.tex
 

\begin{table}[!t]
\renewcommand{\arraystretch}{1.3}
\caption{Statistics of The Experimental Datasets}
\label{table_experiments_data}
\centering
\begin{tabular}{|c||cccc|}
\hline
\bfseries Dataset & \bfseries Node Type & \bfseries Edge Type & \makecell[c]{\bfseries Num. of \\ \bfseries  Nodes } & \makecell[c]{\bfseries Num. of\\ \bfseries Edges } \\
\hline \hline
\makecell[c]{\bfseries Douban\\ \bfseries Movie} & U,M,D,G,A & \makecell[c]{U-U, U-G, \\ U-M, M-A,\\ M-D} &37,557 & 1,687,258\\ \hline
\bfseries DBLP & P,A,V,T & \makecell[c]{P-A, P-V, \\ P-T} & 37,791 & 170,794\\ \hline
\bfseries Cora & P,A,T   & \makecell[c]{P-P, P-T, \\ P-A} & 49,120 & 241,102\\ \hline
\bfseries IMDB & U,M,A,D,G &\makecell[c]{M-U, M-A, \\ M-D, M-G} & 45,519 & 139,741\\ \hline
\bfseries Yelp & Ca,Ci,U,B & \makecell[c]{Ca-B, Ci-B,\\ B-U, U-U} & 324,686 & 3,760,701\\
\hline
\end{tabular}
\end{table}
\section{Experiments}
\label{sec:experiments}
\subsection{Datasets}
\label{Datasets}
We use five publicly available real-world datasets: Douban Movie, DBLP, Cora\footnote{https://people.cs.umass.edu/~mccallum/code-data.html}, IMDB and Yelp to form HIN in the experiments. The detailed description of those datasets is shown in Table \ref{table_experiments_data} and all edges in these datasets are considered as undirected edges.
\textbf{DBLP} and \textbf{Cora} are bibliographical networks. DBLP used here is DBLP-four-areas\cite{reference:dblp_Data}. It is extracted from DBLP database which contains papers(P) that are published in 20 conferences(V)
as well as the related authors(A) and the key words(T) of these papers. As for Cora, there are only 3 types of nodes in this HIN: authors(A), papers(P) and the frequent terms(T) of these papers. Labels of these papers in cora are given based on the research areas of these papers and there are 10 research areas\footnote{Information Retrieval, Databases, Artificial Intelligence, Encryption and Compression, Operating Systems, Networking, Hardware and Architecture, Programming, Data Structures Algorithms and Theory, Human Computer Interaction.} in all. \textbf{IMDB}\cite{reference:shi2018aspem} links the movie-attribute information from IMDB and the user-reviewing information from MovieLens100K\cite{reference:harper2016movielens} to from HIN and there are 5 types of nodes(Movie(M), User(U), Director(D), Genre(G), Actor(A)) in IMDB. Another movie related dataset is \textbf{Douban Movie} which also contains 5 types of nodes(Movie(M), User(U), Director(D), Actor(A), Group(G)) which includes more users interaction information. \textbf{Yelp} is a social media dataset and it is released in Yelp Dataset Challenge. We extract the data of 10 top cities\footnote{the 10 top cities are 'Las Vegas', 'Toronto', 'Phoenix', 'Charlotte', 'Scottsdale', 'Calgary', 'Pittsburgh', 'Montreal', 'Mesa', 'Henderson'} with the most businesses(B) to form the HIN and 4 types of nodes (business(B), user(U), Category of business(Ca), City of business(Ci)) are included in this dataset.
\subsection{Methods to Compare}
\label{subsec:Methods_to_Comapre}
To provide comprehensive evaluation, we tested a number of state-of-the-art methods such as GraphSage\cite{reference:hamilton2017inductive}, HeGAN\cite{reference:hu2019HeGAN} as well as polysemous embedding\cite{reference:liu2019single} and we choose to report the most relevant 5 algorithms which are DeepWalk, LINE, Metapath2vec, AspEm and HIN2vec. We evaluate the effectiveness of the proposed mSHINE against these 5 baseline representation learning methods on two HIN analysis tasks: nodes classification and link prediction. Here, DeepWalk and LINE are originally designed to process homogeneous information networks. In our experiments, these two methods are applied by ignoring the node and edge types in HIN.

\textbf{DeepWalk}\cite{reference:perozzi2014deepwalk} is the first method that applied NLP algorithms(Word2vec\cite{reference:mikolov2013distributed}) in information network representation learning tasks. Specifically, a number of paths are extracted started from each node through random walk, then each path is considered as a sentence and each node is considered as a word. Word2vec then is applied to learning the representation of each node.

\textbf{LINE}\cite{reference:Tang2015LINE} is proposed to preserve both \textit{local} and \textit{global} network structure information through a carefully designed objective function. Different from DeepWalk, the training procedure of LINE is based on edge-sampling and this is more efficient than random-walk-based training which is used in DeepWalk and random-walk-related algorithms.

\textbf{Metapath2vec}\cite{reference:metapath2vec} is proposed based on DeepWalk  and mainly designed for HIN representation learning. There are two main differences compared with DeepWalk: 1) random walk is conducted under the direction of Metapath in Metapath2vec; 2) Metapath2vec applies heterogeneous softmax(negtive sampling) at the prediction layer.

\textbf{AspEm}\cite{reference:shi2018aspem} proposed a new coefficient to measure the incompatibility of different types of edges in an HIN. When the incompatibility of a group of edge types is smaller than a pre-defined threshold, this group of edge types will be considered as an aspect of the HIN and then node representations will be learned within different aspects separately. AspEm does not specify the representation learning algorithm in the original paper, we choose to use the PTE\cite{reference:tang2015pte} in our experiments which is also suggested by AspEm. 

\textbf{HIN2vec}\cite{reference:fu2017hin2vec} proposed to limit the length of possible meta-paths and then learn node and meta-path representations. In this algorithm, a number of paths are extracted through random walk first and then node pairs within a predefined window size as well as the meta-path type that connects these two nodes are taken as a training sample. The learning structure of HIN2vec is designed to predict the existing of a specific type of connection between two nodes.

The selected 5 baselines have a comprehensive coverage of existing HIN representation learning methods. DeepWalk and LINE are the earliest and classic homogeneous information network representation learning methods. Metapath2vec, AspEm and HIN2vec are the typical HIN representation learning methods where Metapath2vec and HIN2vec make use of meta-paths to distinguish different semantic information of HIN while AspEm extracts different aspects by analyzing the structural information of HIN to address the problem of incompatibility. 

\subsection{Implementation Details}
    The implementation details of mSHINE in the experiments are shown as below which is applied on all the datasets:
    \begin{enumerate}[leftmargin=*]
        \item The number initial meta-paths we selected based on the criteria proposed in Section \ref{subsect:Metapath preparation} for all experimental datasets are summarized in Table \ref{table_num_metapaths}.
        \item As for meta-path-related functions $f_{x}^{M_i}$, $f_{h}^{M_i}$ and $f_{y}^{M_i}$, we choose to use Hadamard product in this work:
            \begin{equation}
                 \Vec{x}_{V_n}^{M_i} = f_{x}^{M_i}(\Vec{x}_{V_n}) = \Vec{x}_{V_n} \circ \Vec{v}_{x}^{M_i}
            \end{equation}
            \begin{equation}
                 \Vec{h}_{V_n}^{M_i} = f_{h}^{M_i}(\Vec{h}_{V_n}) = \Vec{h}_{V_n} \circ \Vec{v}_{h}^{M_i}
            \end{equation}
            \begin{equation}
                 W_{hy(V_n)}^{M_i} = f_{y}^{M_i}(W_{hy(V_n)}) = W_{hy(V_n)} \circ \Vec{v}_{y}^{M_i}
            \end{equation}
        where $\Vec{v}_{h}^{M_i}, \Vec{v}_{x}^{M_i}, \Vec{v}_{y}^{M_i} \in R^{d \times 1}$ are trainable vectors. Other meta-path-related functions can also be used while we just take Hadamard product as an example here.
        \item Batch size is set as $B=30$ for each iteration and each training sample should provide a three-element node sequence as well as the corresponding meta-path type.
        \item The dimension of node representations($d$) and negative sampling rate($K$) are set as $128$ and $5$ respectively. Other algorithms will also follow this setting.
        \item The number of epochs is set as $E=1000$ for all datasets.
        \item Stochastic gradient descent (SGD)\cite{reference:bottou1991stochastic} is used to train all the parameters in mSHINE.
        \item When training finished, the learned basic meta-path-related node representations are used in experimental studies to show the performance of mSHINE\footnote{Before training, we use normal distribution to initialize the basic node representations where \textit{mean}$=0$ and \textit{stddev}$=0.1$. The state and output node representations are initialized as $0$. }.
        \item Other hyper-parameters are set as the same as other baseline algorithms which will be introduced individually in each experiment section.
        \item All the experiments are conducted in Nvidia Tesla P100 Cluster.
        \end{enumerate}
        
\begin{table}[t]
\renewcommand{\arraystretch}{1.3}
\caption{The Number of Initial Meta-paths for Each Dataset }
\label{table_num_metapaths}
\centering
\begin{tabular}{|c|ccccc|}
\hline
\bfseries Dataset &
\makecell[c]{\bfseries Douban \\ \bfseries Movie} &
\bfseries DBLP &
\bfseries Cora &
\bfseries IMDB &
\bfseries Yelp\\
\hline
\makecell[c]{\bfseries Num. of\\ \bfseries Meta-paths} &
15&
6&
6&
10&
10\\ \hline

\end{tabular}
\end{table}

\subsection{Node Classification}
    In this section, the effectiveness of mSHINE is evaluated through node classification. Related experimental setup as well as the procedure of constructing classification datasets will be introduced first and then the experimental results will be discussed.
    \subsubsection{Experimental setup}
    
    \begin{table*}
    \centering
    \caption{Performance of Node Classification}
    \label{tab:classification}
    \begin{tabular}{|c|c|c|c|c|c|c|c|c|}
        \hline
        Dataset & Metric & Training & DeepWalk  & LINE & Metapath2vec & AspEm & HIN2vec & mSHINE\\\hline
        \multicolumn{3}{|c|}{Selected Aspect}   &          &     & PPP & CONCAT &     & APPA\\\hline
        \multirow{8}{*}{Cora} & \multirow{4}{*}{f1-macro} & \multicolumn{1}{c|}{20\%} &$0.7267$&$0.7304^*$&$0.7227$& $0.6359$ &$0.7172$& $\textbf{0.7418}$\\
                              &                           & \multicolumn{1}{c|}{40\%} &$0.7577^*$&$0.7573$&$0.7511$& $0.6916$ &$0.7451$& $\textbf{0.7662}$\\
                              &                           & \multicolumn{1}{c|}{60\%} &$0.7759^*$&$0.7671$&$0.7645$& $0.7154$ &$0.7557$& $\textbf{0.7798}$\\
                              &                           & \multicolumn{1}{c|}{80\%} &$0.7811^*$&$0.7793$&$0.7712$& $0.7315$ &$0.7621$& $\textbf{0.7903}$\\
                              \cline{2-9}
                              & \multirow{4}{*}{f1-micro} & \multicolumn{1}{c|}{20\%} &$0.7899$&$0.7928*$&$0.7875$& $0.7304$ &$0.7826$& $\textbf{0.8023}$\\
                              &                           & \multicolumn{1}{c|}{40\%} &$0.8118^*$&$0.8109$&$0.8081$& $0.7620$ &$0.8012$& $\textbf{0.8202}$\\
                              &                           & \multicolumn{1}{c|}{60\%} &$0.8249*$&$0.8184$&$0.8185$& $0.7811$ &$0.8093$& $\textbf{0.8304}$\\
                              &                           & \multicolumn{1}{c|}{80\%} &$0.8318*$&$0.8277$&$0.8235$& $0.7908$ &$0.8148$& $\textbf{0.8378}$\\
                              \hline
        \multicolumn{3}{|c|}{Selected Aspect}   &          &     & BCaB & CONCAT &     & CaBUBCa\\\hline
        \multirow{8}{*}{YELP} & \multirow{4}{*}{f1-macro} & \multicolumn{1}{c|}{20\%} &$0.4867$ &$0.2196$&$0.5100^*$&$0.4763$& $0.4112$ &$\textbf{0.5563}$\\
                              &                           & \multicolumn{1}{c|}{40\%} &$0.5231$ &$0.2491$&$0.5247$&$0.5295^*$& $0.4633$ &$\textbf{0.5840}$\\
                              &                           & \multicolumn{1}{c|}{60\%} &$0.5503^*$ &$0.2575$&$0.5275$&$0.5465$& $0.4876$ &$\textbf{0.5975}$\\
                              &                           & \multicolumn{1}{c|}{80\%} &$0.5566$ &$0.2632$&$0.5360$&$0.5601^*$& $0.5035$ &$\textbf{0.6123}$\\
                              \cline{2-9}
                              & \multirow{4}{*}{f1-micro} & \multicolumn{1}{c|}{20\%} &$0.6215$ &$0.4269$&$0.6736^*$&$0.6543$& $0.6017$&$\textbf{0.6933}$\\
                              &                           & \multicolumn{1}{c|}{40\%} &$0.6428$ &$0.4484$&$0.6790$&$0.6850^*$& $0.6409$&$\textbf{0.7066}$\\
                              &                           & \multicolumn{1}{c|}{60\%} &$0.6604$ &$0.4651$&$0.6830$&$0.6963^*$& $0.6539$&$\textbf{0.7142}$\\
                              &                           & \multicolumn{1}{c|}{80\%} &$0.6646$ &$0.4622$&$0.6834$&$0.7058^*$& $0.6682$&$\textbf{0.7233}$\\
                              \hline
        \multicolumn{3}{|c|}{Selected Aspect}   &          &     & MUM & CONCAT &     &MUGUM \\\hline
        \multirow{8}{*}{Douban Movie} & \multirow{4}{*}{f1-macro} 
                                                          & \multicolumn{1}{c|}{20\%} &$0.0789$&$0.0825^*$&$0.0612$&$0.0811$ &$0.0721$&$\textbf{0.1016}$\\
                              &                           & \multicolumn{1}{c|}{40\%} &$0.1070^*$&$0.0926$&$0.0649$&$0.1002$ &$0.0831$&$\textbf{0.1239}$\\
                              &                           & \multicolumn{1}{c|}{60\%} &$0.1186^*$&$0.1046$&$0.0734$&$0.1098$ &$0.0900$&$\textbf{0.1352}$\\
                              &                           & \multicolumn{1}{c|}{80\%} &$0.1231^*$&$0.1111$&$0.0752$&$0.1175$ &$0.0971$&$\textbf{0.1448}$\\ 
                              \cline{2-9}
                              & \multirow{4}{*}{f1-micro} & \multicolumn{1}{c|}{20\%} &$0.4195$&$0.4575^*$&$0.3944$&$0.4432$ &$0.4397$&$\textbf{0.4922}$\\
                              &                           & \multicolumn{1}{c|}{40\%} &$0.5048^*$&$0.4645$&$0.4007$&$0.4830$ &$0.4651$&$\textbf{0.5157}$\\
                              &                           & \multicolumn{1}{c|}{60\%} &$0.5174^*$&$0.4820$&$0.4298$&$0.4931$ &$0.4778$&$\textbf{0.5249}$\\
                              &                           & \multicolumn{1}{c|}{80\%} &$0.5196^*$&$0.4872$&$0.4361$&$0.5013$ &$0.4856$&$\textbf{0.5349}$\\
                              \hline                      
    \end{tabular}
    \end{table*}
    \textbf{Data preparation}. To evaluate the performance of algorithms through node classification task, HIN datasets which are used for node classification should provide node labels, so, Cora, Yelp and Douban Movie are used as the experimental datasets in this experiment. As introduced in Section \ref{Datasets}, papers in Cora can be categorized by 10 different research areas and these areas are used as paper labels for paper node classification(The property of DBLP is similar to Cora and we choose to use Cora because of its larger dataset size). Following HIN2vec\cite{reference:fu2017hin2vec}, in Yelp, we select 10 main cuisines\footnote{American, French, Italian, Mexican, Canadian, Thai, Indian, Japanese, Chinese, Vietnamese} in restaurants’ categories as labels and there are 17,782 restaurant business in all can be labeled with one of the 10 cuisine types. Besides, we use Douban Movie dataset to evaluate the quality of the learned node representations in multi-label classification task where movie types will be used as movie labels. Thus, the labeled node classification datasets are constructed.
    
    \textbf{Hyper-parameter setting}. For random-walk-based baseline algorithms, for fairly comparing their performance, the number of node sequences that are extracted from each node is set as the same which is $w_n=80$ and the sequence length is set as $w_l=40$ which are the same as DeepWalk paper. As for the window size, we use $w_s=4$ which follows the experimental setting of HIN2vec. Regarding other baseline algorithms, we follow the default setup as what is reported in the original papers. In AspEm, we follow the steps they proposed to select aspects and separately learn node representations for each aspect. At last, we report the best results after comparing all the selected aspects. As for Metapath2vec which requires us manually selecting meta-paths, we only test a limited number of meta-paths for each dataset since it is not practical to test all possible meta-paths. The way we set meta-paths for Metapath2vec follows two steps:
     \begin{inparaenum}[1)]
        \item decide key node types. For example, the key node type in Cora should be $Paper$ since the node classification is conducted based on paper nodes.
        \item list all symmetric meta-paths which start from the key node type and end when the key node type appears at the first time. For example, the tested meta-paths of Cora are $P$$\xrightarrow{}$$A$$\xrightarrow{}$$P$,$P$$\xrightarrow{}$$T$$\xrightarrow{}$$P$ and $P$$\xrightarrow{}$$P$$\xrightarrow{}$$P$.
    \end{inparaenum}
    The same as AspEm, when we report the results of Metapath2vec and mSHINE, we report the best results after comparing all tested meta-paths in these two methods respectively. Regarding HIN2vec, as introduced in Section \ref{subsec:Methods_to_Comapre}, all meta-paths whose length are shorter than the window size $w_s=4$ will be learned. In this case, the number of selected meta-paths(aspects) in Cora, Douban Movie and Yelp is $80$, $176$ and $109$ respectively and it is impossible to separately do classification experiments for each of the meta-paths(aspects). In this experiment, we choose to follow the original setting of HIN2vec where the general node representations are used to conduct node classification task.
    
    Overall, we use the learned node representations from different algorithms as node features and then do node classification with off-the-shelf classifier(support vector machine(SVM) is used in our experiment). The classification performance is evaluated using \textit{f1-macro} and \textit{f1-micro} score respectively\cite{reference:cui2019survey}. For all datasets, following the experimental setting of \cite{reference:Shi2019Herec,reference:Wang2019HAN}, we set four training ratios as in $\{20\%,40\%,60\%,80\%\}$. For each ratio, we randomly generate ten evaluation sets where the learned node representations are divided into training data and test data based on the training ratio and the average of these ten evaluation results on the test data will be reported as the final results.
        
    \subsubsection{Experimental results and discussion}
    \label{subsubsec:node classification result}
    The node classification results are summarized in Table \ref{tab:classification}. The best performance of the existing algorithms are marked with * and the best results for each dataset are highlighted in bold. 
    
    The proposed mSHINE outperforms all the state-of-the-art algorithms in node classification task which illustrates the effectiveness of the proposed method from two angles: the capability of selecting suitable meta-paths(aspects) and the capability of capturing long-term information. 
    
    \textbf{Meta-path(aspect) Selection}. It is easy to find from Table \ref{tab:classification} that when the test data is Yelp, HIN-based algorithms namely Metapath2vec, AspEm, HIN2vec as well as mSHINE are generally better than homogeneous-information-network-based algorithms such as DeepWalk and LINE which indicates the necessary of meta-paths(aspects) selection in Yelp node classification. So we focus on the node classification performance on Yelp to measure the capability of meta-paths(aspects) selection of an algorithm. Table \ref{tab:classification} shows that a worse performance of HIN2vec compared with DeepWalk, which is expected from our understanding of HIN2vec. This algorithm simultaneously learns node and meta-path representations and mainly makes use of meta-path representations to distinguish different types of links or paths. However, from the node representation angle, all semantic information is still mixed within a single node embedding which might still introduce noise. Regarding Metapath2vec and AspEm, they are capable of embedding different semantic information into different node representations. However, for each meta-path(aspect), Metapath2vec and AspEm require re-training the entire model which is extremely inefficient. Therefore, an effective meta-path(aspect) selection strategy is necessary for these two algorithms to reduce training cost while it is still unsolved so far or the proposed strategy is sub-optimal in their work. We test more meta-paths(aspects) or meta-path(aspect) combination of Metapath2vec and AspEm to show the impact of meta-path(aspect) selection strategy and the experimental results are shown in Table \ref{tab:classification_yelp}. Cora and Douban Movie also show the similar phenomenon as Yelp while we only display the results of Yelp here since it is the most obvious one. When Metapath2vec uses the combination of meta-paths $B$$\xrightarrow{}$$U$$\xrightarrow{}$$B$, $B$$\xrightarrow{}$$C_a$$\xrightarrow{}$$B$ and $B$$\xrightarrow{}$$C_i$$\xrightarrow{}$$B$, AspEm uses the combination of aspects $B-U$,$U-U$ and $B-C_a$, these two algorithms reach comparable performance to mSHINE where only one meta-path($C_a$$\xrightarrow{}$$B$$\xrightarrow{}$$U$$\xrightarrow{}$$B$$\xrightarrow{}$$C_a$) is used to learn node representations. However, these effective meta-paths(aspects) combination shown in Table \ref{tab:classification_yelp} cannot be easily detected through their original method while mSHINE is able to address this problem in a feasible way. mSHINE learns multiple node representations from multiple meta-paths simultaneously, and then users can select the most suitable meta-path and node representations based on specific classification tasks, without extra efforts after training once.
    
    \begin{table*}
    \centering
    \caption{Performance of Node Classification with Cora with Various Window Size from $w_s=1$ to $w_s=3$}
    \label{tab:classification_cora}
    \begin{tabular}{|c|c|c|c|c|c|c|c|c|c|c|}
        \hline
        Metric & Training                   & \multicolumn{3}{c|}{DeepWalk}  &\multicolumn{3}{c|}{Metapath2vec} & \multicolumn{3}{c|}{HIN2vec}\\\hline
        \multicolumn{2}{|c|}{Selected Aspects}  &\multicolumn{3}{c|}{} & \multicolumn{3}{c|}{PPP}&\multicolumn{3}{c|}{} \\\hline
        \multicolumn{2}{|c|}{Window Size}&$w_s=1$ & $w_s=2$  & $w_s=3$ & $w_s=1$  & $w_s=2$  & $w_s=3$& $w_s=1$  & $w_s=2$  & $w_s=3$ \\\hline
        \multirow{4}{*}{f1-macro} & \multicolumn{1}{c|}{20\%} &$0.3946$&$0.6845$&$\textbf{0.7113}$& $0.5136$ &$0.6411$& $\textbf{0.6918}$& $0.7088$ &$0.6870$& $\textbf{0.7154}$\\
                                  & \multicolumn{1}{c|}{40\%} &$0.4933$&$0.7306$&$\textbf{0.7491}$& $0.6255$ &$0.6928$& $\textbf{0.7234}$& $0.7367$ &$0.7281$& $\textbf{0.7445}$\\
                                  & \multicolumn{1}{c|}{60\%} &$0.5401$&$0.7482$&$\textbf{0.7679}$& $0.6696$ &$0.7109$& $\textbf{0.7387}$& $0.7495$ &$0.7481$& $\textbf{0.7540}$\\
                                  & \multicolumn{1}{c|}{80\%} &$0.5717$&$0.7577$&$\textbf{0.7743}$& $0.6874$ &$0.7239$& $\textbf{0.7504}$& $0.7567$ &$0.7542$& $\textbf{0.7656}$\\
                              \cline{1-11}
        \multirow{4}{*}{f1-micro} & \multicolumn{1}{c|}{20\%} &$0.6017$&$0.7592$&$\textbf{0.7774}$& $0.6255$ &$0.7206$& $\textbf{0.7623}$& $0.7728$ &$0.7635$& $ \textbf{0.7793}$\\
                                  & \multicolumn{1}{c|}{40\%} &$0.6470$&$0.7900$&$\textbf{0.8040}$& $0.7037$ &$0.7555$& $\textbf{0.7821}$& $ 0.7937$ &$0.7877$& $\textbf{0.7978}$\\
                                  & \multicolumn{1}{c|}{60\%} &$0.6681$&$0.8043$&$\textbf{0.8171}$& $0.7354$ &$0.7703$& $\textbf{0.7955}$& $0.8025$ &$0.8005$& $\textbf{0.8050}$\\
                                  & \multicolumn{1}{c|}{80\%} &$0.6830$&$0.8090$&$\textbf{0.8238}$& $0.7555$ &$0.7800$& $\textbf{0.8021}$& $0.8092$ &$0.8050$& $\textbf{0.8136}$\\
                              \hline
        \end{tabular}
    \end{table*}
    \textbf{Long-term information}. From data Cora, it is easy to find that the performance of long-term-information-based algorithms such as DeepWalk, HIN2vec, Metapath2vec as well as mSHINE are generally better than short-term-information-based algorithms(AspEm) which indicates the importance of long-term information in Cora node classification. So the better performance of mSHINE in Cora justifies its effectiveness in capturing long-term information. In DeepWalk, HIN2vec and Metapath2vec, a larger window size indicates a longer long-term information the algorithms are capturing. The influence of window size to the classification performance on Cora is shown in Table \ref{tab:classification_cora}. We highlight the best window size for each algorithm where we only compare the results of window size ranges from $w_s=1$ to $w_s=3$ ($w_s=4$ is illustrated in Table \ref{tab:classification}) to illustrate the trend. Basically, the trend can also be found in other datasets while only the results on Cora are reported since it is the most representative one in this experiment. It is obvious that as the window size increases, the classification performance becomes better for Metapath2vec and DeepWalk. As for HIN2vec, increasing window size also leads to the increasing of meta-path types, so the result does not show a monotonous growth. By displaying the impact of window size on the classification performance, we could infer the importance of long-term information in Cora node classification, which further infers the capability of mSHINE in capturing long-term information.
    
    In summary, the property of capturing long-term information in DeepWalk, HIN2vec and Metapath2vec is controlled by window size($w_s$) and the ability of meta-path(aspect) selection is controlled by meta-path length(HIN2vec), incompatibility threshold(AspEm) or even no criteria(Metapath2vec). Compared with them, mSHINE does not require these hyper-parameters which makes mSHINE more feasible for practical usage besides its state-of-the-art performance as shown in Table \ref{tab:classification}. 
    
\subsection{Link Prediction}
    In this section, mSHINE is evaluated via another HIN analysis task which is link prediction. In this experiment, the link prediction is modeled as ranking all possible nodes based on a query node and the performance of link prediction is evaluated through the relevance between the highly ranked nodes and the query node. The related experimental setup will be introduced first which is followed by experimental results and the detailed analysis.
    \subsubsection{Experimental setup}
        \textbf{Data preparation}. In this experiment, we adopt three widely used datasets from different domains as the experimental datasets which are IMDB, Yelp and DBLP. Following \cite{reference:fu2017hin2vec}, to construct link prediction datasets, for each dataset, we randomly remove $20\%$ of a certain type of edges and the edge types which are selected to remove in DBLP, IMDB and Yelp are set as $P$$-$$A$, $U$$-$$M$ and $B$$-$$U$ respectively. Thus, the experimental datasets are constructed. 
        
    \begin{table*}
    \centering
    \caption{Performance of Node Classification with Yelp with various Meta-paths(Aspects)}
    \label{tab:classification_yelp}
    \begin{tabular}{|c|c|c|c|c|c|c|c|}
        \hline
        Metric & Training                   & \multicolumn{3}{c|}{Metapath2vec}  &\multicolumn{3}{c|}{AspEm} \\\hline
        \multicolumn{2}{|c|}{Selected Aspect}&BUB & BCiB & \makecell[c]{ BUB+BCiB+BCaB  }& BCa+BCi & BU+UU & BU+UU+BCa\\\hline
        \multirow{4}{*}{f1-macro} & \multicolumn{1}{c|}{20\%} &$0.2765$&$0.1096$&$\textbf{0.5575}$& $\textbf{0.5378}$ &$0.1995$& $0.5208$\\
                                  & \multicolumn{1}{c|}{40\%} &$0.3110$&$0.1060$&$\textbf{0.5904}$& $0.5536$ &$0.2189$& $\textbf{ 0.5593}$\\
                                  & \multicolumn{1}{c|}{60\%} &$0.3385$&$0.1022$&$\textbf{0.6085}$& $0.5616$ &$0.2360$& $\textbf{0.5837}$\\
                                  & \multicolumn{1}{c|}{80\%} &$0.3440$&$0.1071$&$\textbf{0.6183}$& $0.5628$ &$0.2478$& $\textbf{0.5970}$\\
                              \cline{1-8}
        \multirow{4}{*}{f1-micro} & \multicolumn{1}{c|}{20\%} &$0.4095$&$0.3070$&$\textbf{0.7035}$& $\textbf{0.6847}$ &$0.3699$& $0.6801$\\
                                  & \multicolumn{1}{c|}{40\%} &$0.4309$&$0.3084$&$\textbf{0.7215}$& $0.6961$ &$0.3801$& $\textbf{0.7020}$\\
                                  & \multicolumn{1}{c|}{60\%} &$0.4442$&$0.3089$&$\textbf{0.7305}$& $0.6992$ &$0.3872$& $\textbf{0.7131}$\\
                                  & \multicolumn{1}{c|}{80\%} &$0.4485$&$0.3105$&$\textbf{0.7352}$& $0.6997$ &$0.3989$& $\textbf{0.7182}$\\
                              \hline
        \end{tabular}
    \end{table*}
        \textbf{Evaluation metrics}. After removing edges, network representation learning algorithms are applied to learn HIN node representations based on the remained HIN. The purpose of link prediction is to reconstruct those removed edges. In mSHINE, the prediction objective function given in Eq. \ref{equ:objective_1}, can be used to rank all possible nodes based on the prediction value. Here, all possible nodes means the nodes with the same type as the target nodes. Since mSHINE is capable of simultaneously capturing information learned from different meta-paths, we will calculate the average relevance based on multiple meta-paths and use this average value to rank nodes. For example, in DBLP, to predict the related authors($A$) of a given paper($P$), we will calculate the prediction value based on different meta-paths and different three-element node sequence including $V$$-$$P$$\xrightarrow{}$$A$, $T$$-$$P$$\xrightarrow{}$$A$ and $A$$-$$P$$\xrightarrow{}$$A$, then the average value from these meta-paths is used to rank all authors in the HIN. In this case, the well trained node representation learning module which is introduced in Section \ref{subsect:Rpresentation Learning} can be easily applied in this link prediction task. Top-k recall, top-k precision, Mean Average Precision(MAP) as well as Mean reciprocal rank (MRR) are chosen as metrics for link prediction evaluation\cite{reference:cui2019survey} here and the four metrics are given below.
        \begin{enumerate}[leftmargin=*]
            \item \textit{top-k precision}:
            \begin{displaymath}
                Pre@k = \frac{\quad num.\quad of\quad hits\quad @k \quad}{k}
            \end{displaymath}
            \item \textit{top-k recall}:
            \begin{displaymath}
                Rec@k = \frac{\quad num.\quad of\quad hits\quad @k \quad}{\quad num.\quad of\quad relevant \quad nodes\quad}
            \end{displaymath}
            \item \textit{MAP}:
            \begin{displaymath}
                AP@k(V_i) = \frac{1}{k}\sum_{j=1}^k Pre@j \cdot relevant@j\\
            \end{displaymath}
            where $relevant@j=1$ if the $j$th recommended node is relevant otherwise $relevant@j=0$ and:
            \begin{displaymath}
                MAP=\frac{1}{N}\sum_{i=1}^NAP@k(V_i)
            \end{displaymath}
            where $N$ represents the number of all query nodes.
            \item \textit{MRR}:
            \begin{displaymath}
                MRR = \frac{1}{N}\sum_{i=1}^N \frac{1}{rank_{V_i}}
            \end{displaymath}
            where $rank_{V_i}$ refers to the rank position of the first relevant node in recommendation list when the query node is $V_i$. MAP gives a general measure of the link prediction quality while MRR only cares about the highest-ranked relevant item. 
        \end{enumerate}
        As for other baseline algorithms, we also make use of the objective function proposed in respective paper to measure the relevance between nodes. It is worth noting that the objective function value of HIN2vec is calculated based on specific meta-paths and this requires us specifying the meta-path type used for conducting link prediction. So, we choose to use the same meta-path as the removed edge type for each dataset. For example, the removed edge type in DBLP is $P$$-$$A$, so the meta-path type we choose to measure the relevance between nodes in DBLP for HIN2vec is $P$$-$$A$.
    
    \begin{table*}
    \centering
    \caption{Performance of Link Prediction}
    \label{tab:link_prediction}
    \begin{tabular}{|c|c|c|c|c|c|c|c|c|}
        \hline
        Dataset & \multicolumn{2}{c|}{Metric} & DeepWalk &LINE &Metapath2vec & AspEm & HIN2vec & mSHINE\\\hline
        \multicolumn{3}{|c|}{Selected Aspect}   &          &     & PAP & CONCAT &  PA   & \\\hline
        \multirow{8}{*}{DBLP} & \multirow{4}{*}{Pre@k}  & k=1 &$0.0690$ &$0.0039$&$0.0648$&$0.1694*$&$0.0392$ &$\textbf{0.1867}$\\
                              &                         & k=3 &$0.0438$ &$0.0032$&$0.0610$&$0.1055*$&$0.0326$ &$\textbf{0.1100}$\\
                              &                         & k=10&$0.0221$ &$0.0024$&$0.0429$&$0.0496*$&$0.0200$ &$\textbf{0.0501}$\\
                              \cline{2-9}
                              & \multirow{4}{*}{Rec@k}  & k=1 &$0.0587$ &$0.0033$&$0.0553$&$0.1483*$&$0.0342$ &$\textbf{0.1620}$\\
                              &                         & k=3 &$0.1095$ &$0.0080$&$0.1573$&$0.2702*$&$0.0824$ &$\textbf{0.2791}$\\
                              &                         & k=10&$0.1831$ &$0.0194$&$0.3701$&$\textbf{0.4179}*$&$0.1669$ &$\textbf{0.4179}$\\
                              \cline{2-9}
                              & \multicolumn{2}{c|}{MAP}      &$0.1045$&$0.0112$&$0.1470$&$0.2415*$&$0.0803$ &$\textbf{0.2550}$\\
                              & \multicolumn{2}{c|}{MRR}      &$0.1149$&$0.0130$&$0.1152$&$0.2577*$&$0.0870$ &$\textbf{0.2763}$\\
                              \hline
        \multicolumn{3}{|c|}{Selected Aspect}   &          &     & UMAMU & CONCAT &  UM  & \\\hline
        \multirow{8}{*}{IMDB} & \multirow{4}{*}{Pre@k}  & k=1 &$0.0021$&$0.0043$&$0.0224$&$0.2145$&$\textbf{0.3202}*$ &$0.3180$\\
                              &                         & k=3 &$0.0025$&$0.0039$&$0.0164$&$0.2166$&$0.2832*$ &$\textbf{0.2942}$\\
                              &                         & k=10&$0.0038$&$0.0041$&$0.0161$&$0.1892$&$\textbf{0.2344}*$ &$0.2334$\\
                              \cline{2-9}
                              & \multirow{4}{*}{Rec@k}  & k=1 &$0.0001$&$0.0002$&$0.0025$&$0.0222$&$0.0335*$ &$\textbf{0.0370}$\\
                              &                         & k=3 &$0.0003$&$0.0006$&$0.0093$&$0.0644$&$0.0867*$ &$\textbf{0.0917}$\\
                              &                         & k=10&$0.0031$&$0.0027$&$0.0288$&$0.1729$&$\textbf{0.2257}*$ &$0.2186$\\
                              \cline{2-9}
                              &\multicolumn{2}{c|}{MAP}      &$0.0229$&$0.0126$&$0.0249$&$0.1720$&$0.2195*$ &$\textbf{0.2212}$\\
                              &\multicolumn{2}{c|}{MRR}      &$0.0253$&$0.0226$&$0.0405$&$0.3984$&$\textbf{0.4981}*$ &$\textbf{0.4981}$\\
                              \hline
        \multicolumn{3}{|c|}{Selected Aspect}   &          &     & UBCiBU & CONCAT & BU  & \\\hline
        \multirow{8}{*}{YELP} & \multirow{4}{*}{Pre@k}  & k=1 &$0.0033$&$0.0072$&$0.0145*$&$0.0134$&$0.0048$ &$\textbf{0.0180}$\\
                              &                         & k=3 &$0.0028$&$0.0066$&$0.0116*$&$0.0111$&$0.0039$ &$\textbf{0.0158}$\\
                              &                         & k=10&$0.0023$&$0.0041$&$0.0079$&$0.0087*$&$0.0031$ &$\textbf{0.0131}$\\
                              \cline{2-9}
                              & \multirow{4}{*}{Rec@k}  & k=1 &$0.0008$&$0.0013$&$0.0005$&$0.0028*$&$0.0018$ &$\textbf{0.0048}$\\
                              &                         & k=3 &$0.0017$&$0.0040$&$0.0012$&$0.0066*$&$0.0040$ &$\textbf{0.0121}$\\
                              &                         & k=10&$0.0040$&$0.0081$&$0.0027$&$0.0155*$&$0.0097$ &$\textbf{0.0297}$\\
                              \cline{2-9}
                              &\multicolumn{2}{c|}{MAP}&$0.0030$&$0.0045$&$0.0022$&$0.0096*$&$0.0077$ &$\textbf{0.0169}$\\
                              &\multicolumn{2}{c|}{MRR}&$0.0083$&$0.0129$&$0.0219$&$0.0248*$&$0.0131$ &$\textbf{0.0354}$\\
                              \hline
    \end{tabular}
    \end{table*}
    \subsubsection{Experimental result and discussion}
    The performance of each algorithm in link prediction task is shown in Table \ref{tab:link_prediction}. The same as Section \ref{subsubsec:node classification result}, the best performance of the existing algorithms are marked with * and we highlight the best result for each dataset in bold. 
    
    \begin{table*}
    \centering
    \caption{Performance of Link Prediction Based on Different Meta-path(Aspects) }
    \label{tab:link_prediction_heterogeity}
    \begin{tabular}{|c|c|c|c|c|c|c|c|}
        \hline
        Dataset & \multicolumn{2}{c|}{Metric} & \multicolumn{3}{c|}{Metapath2vec} & \multicolumn{2}{c|}{HIN2vec} \\\hline
        \multicolumn{3}{|c|}{Selected Aspect}   &    APTPA      & APVPA     & \makecell[c]{APA\\+APTPA+APVP}& PA & \makecell[c]{PA+PTPA\\+PAPA+PVPA}\\\hline
        \multirow{8}{*}{DBLP} & \multirow{4}{*}{Pre@k}  & k=1 &$0.0690$ &$0.0120$&$\textbf{0.1781}$&$0.0392$ &$\textbf{0.1167}$\\
                              &                         & k=3 &$0.0408$ &$0.0083$&$\textbf{0.1173}$&$0.0326$ &$\textbf{0.0671}$\\
                              &                         & k=10&$0.0229$ &$0.0050$&$\textbf{0.0553}$&$0.0200$ &$\textbf{0.0320}$\\
                              \cline{2-8}
                              & \multirow{4}{*}{Rec@k}  & k=1 &$0.0564$ &$0.0103$&$\textbf{0.1563}$&$0.0342$ &$\textbf{0.0982}$\\
                              &                         & k=3 &$0.1059$ &$0.0218$&$\textbf{0.3045}$&$0.0824$ &$\textbf{0.1672}$\\
                              &                         & k=10&$0.1967$ &$0.0447$&$\textbf{0.4749}$&$0.1669$ &$\textbf{0.2634}$\\
                              \cline{2-8}
                              & \multicolumn{2}{c|}{MAP}      &$0.1033$&$0.0229$&$\textbf{0.2600}$&$0.0803$ &$\textbf{0.1581}$\\
                              & \multicolumn{2}{c|}{MRR}      &$0.0751$&$0.0178$&$\textbf{0.1793}$&$0.0870$ &$\textbf{0.1775}$\\
                              \hline
        \multicolumn{3}{|c|}{Selected Aspect}   &     MUM     &   UMDMU  & \makecell[c]{MUM+UMDMU\\+UMGMU+UMAMU} & UM &    \makecell[c]{UM\\+UMUM+UMAM\\+UMGM+UMDM}\\\hline
        \multirow{8}{*}{IMDB} & \multirow{4}{*}{Pre@k}  & k=1 &$0.0149$&$0.0043$&$\textbf{0.0202}$&$\textbf{0.3202}$ &$0.2006$\\
                              &                         & k=3 &$0.0139$&$0.0068$&$\textbf{0.0171}$&$\textbf{0.2832}$ &$0.1572$\\
                              &                         & k=10&$0.0127$&$0.0086$&$\textbf{0.0186}$&$\textbf{0.2344}$ &$0.1249$\\
                              \cline{2-8}
                              & \multirow{4}{*}{Rec@k}  & k=1 &$0.0007$&$0.0022$&$\textbf{0.0032}$&$\textbf{0.0335}$ &$0.0218$\\
                              &                         & k=3 &$0.0032$&$0.0052$&$\textbf{0.0071}$&$\textbf{0.0867}$ &$0.0446$\\
                              &                         & k=10&$0.0181$&$0.0144$&$\textbf{ 0.0318}$&$\textbf{0.2257}$ &$0.1093$\\
                              \cline{2-8}
                              &\multicolumn{2}{c|}{MAP}      &$0.0215$&$0.0223$&$\textbf{0.0298}$&$\textbf{0.2195}$ &$0.1152$\\
                              &\multicolumn{2}{c|}{MRR}      &$0.0359$&$0.0280$&$\textbf{0.0426}$&$\textbf{0.4981}$ &$0.3425$\\
                              \hline
    \end{tabular}
    \end{table*}
    mSHINE outperforms all the state-of-the-art algorithms except HIN2vec when the test data is IMDB. Even comparing with HIN2vec with IMDB, mSHINE achieves top 2 and has comparable performance with HIN2vec. Besides, the effectiveness of the proposed method is quite obvious especially when compared with those algorithms which are designed for the homogeneous information networks and this also indicates the advantages of separately embedding different semantic or structural information in HINs. This is because that the purpose of link prediction is to predict a specific type of edges while algorithms such as DeepWalk and LINE fail to distinguish different edge types and embedding all types of edge information into one embedding might bring in too much noise which is harmful to link prediction performance. However, even different semantic information is separately embedded, the way to select and merge different semantic information learned from different meta-paths(aspects) also influence link prediction performance. We take DBLP and IMDB as the examples to show the influence of meta-paths(aspect) selection in link prediction and the experimental results are shown in Table \ref{tab:link_prediction_heterogeity}. We highlight the best meta-path or meta-path combination for each algorithms in bold. As for Metapath2vec, it usually gets better performance when combining more meta-paths information while the training cost is huge since it requires re-training for each meta-path separately. Regarding HIN2vec, it is capable of simultaneously learning multiple meta-path embeddings which reduces the cost of training. However, combining representations learned from different meta-paths does not ensure that the performance of HIN2vec in link prediction is improved. To obtain the optimal performance of HIN2vec, more meta-path combinations need to be tested while this will increase the cost of testing. On the contrary, we can find from Table \ref{tab:link_prediction} that mSHINE is able to get satisfied performance by directly combining node representations learned from all different meta-paths so that no more meta-path combination testing is required. Again, compared with other baseline algorithms, the proposed method requires a smaller number of hyper-parameters while achieves a better performance on almost all datasets which shows the effectiveness and feasibility of mSHINE in practical applications.

%% file: Sections/Section_Conclusion.tex
\section{Conclusion and Future Work}
\label{sec:conclusion}
In this paper, we presented a novel HIN representation learning framework which consists of a set of criteria for initial meta-path selection and an HIN representation learning module. The proposed criteria for initial meta-path selection is helpful to reduce the cost of meta-path exploration during training. The representation learning module inspired by basic RNN structure is designed to be capable of simultaneously learning multiple node representations for different meta-paths which improves the learning efficiency especially when multiple meta-paths(aspects) need to be trained. Besides, the learned node representation learning module can be easily applied in downstream link prediction tasks. Through our comprehensive experiments on 5 real-world datasets, we demonstrates the capability of mSHINE in both capturing long-term information as well as efficiently selecting suitable meta-paths for downstream applications without any prior knowledge from experts. In these experiments, the proposed method outperforms the stat-of-the-art methods on most of the datasets in both node classification and link prediction tasks. Besides, by storing state node representations during training, compared with other random-walk-based algorithms, the proposed method is able to capture long-term node information without preparing a huge number of long node sequences for training which makes mSHINE more efficient.

As for future work, we will explore more effective meta-path-related functions to improve the interaction between highly-related meta-paths which is helpful for integrating semantic information from different but related meta-paths. Besides, the meta-path selection criteria proposed in this work is only a preliminary criteria for initial meta-path selection. A more automatic and dynamic way to select suitable meta-paths or meta-path combinations may be investigated in the future. Furthermore, except the heterogeneity, real-world networks are often noisy and uncertain\cite{8395024}, which requires a more robust network embedding algorithm to produce stable and robust representations and we will explore a more robust method to improve the performance.